%% file: 0-gecco17-NI.tex
\documentclass[sigconf]{acmart}

\usepackage{array}
\usepackage{booktabs} 
\usepackage{multirow} 
\usepackage{subcaption}
\usepackage{verbatim} 

\newcommand{\etal}[0]{\emph{et al.}}

\newcommand{\down}{{\color[rgb]{0.7,0,0}{$\blacktriangledown$}}}
\newcommand{\up}{{\color[rgb]{0,0.45,0.1}{$\blacktriangle$}}}
\newcommand{\eq}{{\color[rgb]{0,0,0.5}{$\blacklozenge$}}}

\newcolumntype{A}{>{\raggedright\arraybackslash}m{1.65cm}}
\newcolumntype{B}{>{\raggedright\arraybackslash}m{1.75cm}}

\begin{document}

\title{How Noisy Data Affects Geometric Semantic\\Genetic Programming}

\author{Luis F. Miranda}
\email{luisfmiranda@dcc.ufmg.br}

\author{Luiz Otavio V. B. Oliveira}
\email{luizvbo@dcc.ufmg.br}
\affiliation{%
  \institution{Computer Science Department\\Universidade Federal de Minas Gerais}
  \city{Belo Horizonte} 
  \state{Brazil}
}

\author{Joao Francisco B. S. Martins}
\email{joaofbsm@dcc.ufmg.br}

\author{Gisele L. Pappa}
\email{glpappa@dcc.ufmg.br}
\affiliation{%
  \institution{Computer Science Department\\Universidade Federal de Minas Gerais}
  \city{Belo Horizonte} 
  \state{Brazil}
}


\renewcommand{\shortauthors}{Luis F. Miranda et al.}

\begin{abstract}

Noise is a consequence of acquiring and pre-processing data from the environment, and shows fluctuations from different sources---e.g., from sensors, signal processing technology or even human error. As a machine learning technique, Genetic Programming (GP) is not immune to this problem, which the field has frequently addressed. 
Recently, Geometric Semantic Genetic Programming (GSGP), a semantic-aware branch of GP, has shown robustness and high generalization capability. Researchers believe these characteristics may be associated with a lower sensibility to noisy data.
However, there is no systematic study on this matter. This paper performs a deep analysis of the GSGP performance over the presence of noise. Using 15 synthetic datasets where noise can be controlled, we added different ratios of noise to the data and compared the results obtained with those of a canonical GP. The results show that, as we increase the percentage of noisy instances, the generalization performance degradation is more pronounced in GSGP than GP. However, in general, GSGP is more robust to noise than GP in the presence of up to 10\% of noise, and presents no statistical difference for values higher than that in the test bed.


\end{abstract}

\copyrightyear{2017} 
\acmYear{2017} 
\setcopyright{acmlicensed}
\acmConference{GECCO '17}{July 15-19, 2017}{Berlin, Germany}\acmPrice{15.00}\acmDOI{http://dx.doi.org/10.1145/3071178.3071300}
\acmISBN{978-1-4503-4920-8/17/07}

%
%




\maketitle

\input{1-introduction}
\input{2-gsgp}
\input{3-related}

\input{4-methodology}
\input{5-experiments}
\input{6-conclusions}

\begin{acks}
The authors would like to thank the anonymous reviewers for their valuable comments and suggestions. This work was partially supported by the following Brazilian Research Support Agencies: CNPq, FAPEMIG, and CAPES.
\end{acks}

\bibliographystyle{ACM-Reference-Format}
\bibliography{bibliography} 

\end{document}

%% file: 1-introduction.tex
\section{Introduction}

The presence of noise in data is an issue recurrently approached in the machine learning field. Noisy data can highly influence the performance of machine learning techniques, leading to overfitting and poor data generalization \cite{nettleton2010study}. We define noise as anything that obscures the relationship between the predictor variables and the target variable of a problem \cite{hickey1996noise}. In classification and regression problems, noise can be found in the input (predictor) variables, in the output (target) variable or both, and is usually the result of non-systematic errors during the process of data generation.

In the context of regression problems, robust regression methods have been proposed to address noisy data points or outliers\footnote{We consider that both noisy points and outliers are out of pattern instances that should be identified. We do  not go into the merit of whether a noisy point may be actually useful to the task and represent an outlier.}, and also to deal with other data assumptions most regression methods do not respect \cite{rousseeuw2005robust}, such as the independence between the input variables. Although not very popular for some time due to its computational cost, robust regression provide an alternative to deal with noise.
When modeling Genetic Programming (GP) to solve symbolic regression problems, only a few studies have looked at the impact of noise on the results of data generalization and overfitting \cite{borrelli2006performance,sivapragasam2007genetic,defalco2007parsimony,imada2008using}.

Instead, the community has given great focus to the relations between complexity, overfitting and generalization, and its relation to bloat and parsimony \cite{fitzgerald:2014,Vanneschi:2010:bloat}.
These are indeed close-related issues in GP, but they do not account for problems that are not inherent to the GP search, but intrinsic to the input data.
A few works have also investigated this matter considering the behavior of the GP when additive noise is added to the input data \cite{borrelli2006performance,sivapragasam2007genetic,defalco2007parsimony,imada2008using}.

The main objective of this work is not to look at how canonical GP deals with noise, but rather investigate how GPs that take semantics into account deal with the problem when compared to GP. A few papers in the literature have claimed Geometric Semantic Genetic Programming (GSGP) to be more robust to overfitting---which can be caused by noisy data points---when compared to canonical GP techniques \cite{vanneschi2013new,castelli2012efficient,castelli2013efficient,vanneschi2014genetic,vanneschi2014improving}.
At first, this might be even counter-intuitive, as the exponential growth of solutions caused by GSGP might even worsen the effects of overfitting.
However, no systematic study has been performed to assess whether and in which situations this might be true, and how noisy data points affect the performance of GSGP.

We are particularly interested in noise found in the output variable of symbolic regression problems. This is because GSGP operates in a semantic space, guided by the vector of outputs 
defined by the training set. Hence, noise in the output has a much bigger impact in the search process in GSGP than noise in the predicted variables.

In order to investigate the impacts of noise, we systematically introduced additive noise to a set of 15 artificial datasets from the literature. We evaluated how increasing noise affected the results of error of the methods in both the training and test sets, using a total of 165 versions of the datasets. We also adapted two measures from the classification literature that capture the noise robustness of a method, and present results for these measures in the datasets considered.

In general, our results show that, although GSGP performs better in low levels of noise, as we increase the percentage of noisy instances, the performance of GSGP and GP tend to approximate.

%% file: 2-gsgp.tex
\section{Geometric Semantic Genetic Programming}

Previous GP works have shown that the evolutionary search can be improved through the inclusion of semantic information \cite{vanneschi2014survey}. Among them, the Geometric Semantic GP (GSGP) introduces geometric semantic crossover and mutation operators that, acting on the syntax of the parent programs, produce offspring with known semantic properties \cite{Moraglio2012geometric}.

From the symbolic regression perspective, given a training set $T=\{(\mathbf{x}_i,y_i)\}_{i=1}^n$---where $(\mathbf{x}_i,y_i)\in \mathbb{R}^d\times \mathbb{R}$ ($i=1,2,\ldots,n$)---the semantics of an individual representing a program $p$, denoted by $s(p)$, is defined as the vector of outputs it produces when applied to the set of inputs defined by $T$, i.e.,  $s(p)=[p(\mathbf{x}_1), p(\mathbf{x}_2), \ldots, p(\mathbf{x}_n)]$. This definition allows the semantics of any program to be straightforwardly represented in a $n$-dimensional semantic space $\mathcal{S}$, where $n$ is the size of the training set. Notice that the target output vector defined by the training set---given by $t=[y_1, y_2, \ldots, y_n]$---is also representable in $\mathcal{S}$.




The Geometric Semantic Crossover (GSX) operator combines two parent individuals, $p_1$ and $p_2$, generating an offspring placed in the metric segment in $\mathcal{S}$ connecting both parents:

\begin{equation}
	\label{eq:gss}
	GSX(p_1, p_2)= r\cdot p_1+(1-r)\cdot p_2\enspace,
\end{equation} 

\noindent
where $r$ is a random real constant in [0, 1] (for fitness function based on Euclidean distance) or a random real function with codomain $[0,1]$ 
(for fitness function based on Manhattan distance).

The Geometric Semantic Mutation (GSM) operator, in turn, produces semantic perturbations to a given individual $p$, such that its resulting semantics is placed in a ball with radius $\varepsilon$, proportional to the mutation step $ms$, as given by:

\begin{equation}
	\label{eq:gsm}
	GSM(p)=p+ms\cdot(tr_1-tr_2)\enspace,
\end{equation} 

\noindent
where $tr_1$ and $tr_2$ are real functions randomly generated.

\begin{figure}[t]
	\centering
	\begin{subfigure}{0.4\linewidth}
		\includegraphics[width=\linewidth]{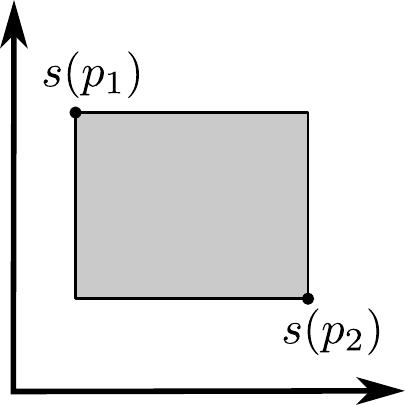}
		\caption{}
		\label{sfig:geo-rep-xover-l1}
	\end{subfigure}
	\hfill
	\begin{subfigure}{0.4\linewidth}
		\includegraphics[width=\linewidth]{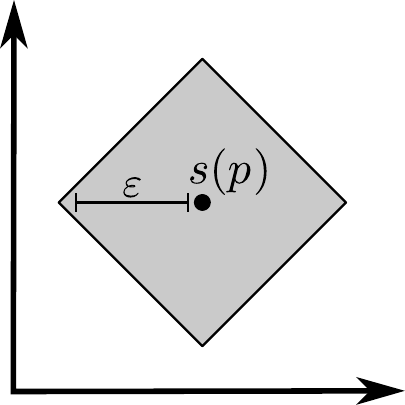}
		\caption{}
		\label{sfig:geo-rep-mutation-l1}
	\end{subfigure}
	\caption{Geometric representation of the geometric semantic (a) crossover and (b) mutation operators in two-dimensional semantic spaces for fitness function based on Manhattan distance.}
	\label{fig:geo-operators}
\end{figure}

Figure \ref{fig:geo-operators} shows the representation of the geometric semantic operators in the semantic space with a Manhattan-based fitness function. The resulting offspring of these operators over the semantics of $p_1$ and $p_2$ are placed in the grey area.

%% file: 3-related.tex
\section{Related Work}


As previously mentioned, the GP community has given a lot of attention to the relations between complexity, overfitting and generalization, and their association to bloat and parsimony \cite{fitzgerald:2014,Vanneschi:2010:bloat}.
This section focuses specifically on works performed to analyze and minimize the effects of noisy data in GP. In addition, to the best of our knowledge, so far there are no measures to quantify the impact of noise in GP-induced models for symbolic regression problems. Thus, we also present an overview of techniques to measure the impact of noisy data on the performance of classification techniques, which we adapted to the regression domain.

\subsection{Genetic Programming with Noisy Data}\label{sec:gp-noisy}

Different strategies have been proposed in symbolic regression to investigate and minimize the impact of noisy data on the search performed by GP. On the one hand, one can try to filter out noise data before performing the regression. On the other hand, one can improve the methods to simply deal with the problem---a much more common approach.

Following the first strategy, 
Sivapragasam \etal{} \cite{sivapragasam2007genetic} use Singular Spectrum Analysis (SSA) to filter out the noise components before performing the symbolic regression of a short time series of fortnight river flow.
The experimental study indicates that when the stochastic (noise) components are removed from short and noisy time-series, the short-lead forecasts can be improved.

Regarding methods that try to deal with the problem, 
Borrelli \etal{} \cite{borrelli2006performance} employ a Pareto multi-objective GP for symbolic regression of time series with additive and multiplicative noise. The authors adopt two different configurations employing statistical metrics for the fitness objectives: (1) the Mean Squared Error (MSE) combined with the first two momenta and (2) the MSE with the skewness added to the kurtosis---all the measures computed regarding the desired and evaluated outputs. An experimental analysis considering time series generated from 50 functions from the literature shows that, although reducing overfitting and bloat, the multi-objective approach does not perform well when the noise level is too high. However, for moderate noise levels, the approach can successfully discover the trend of the series.

De Falco \etal{} \cite{defalco2007parsimony}, in turn, present two GP methods guided by context-free grammars with different fitness functions that take parsimony and the simplicity of the solutions into account. The Parsimony-based Fitness Algorithm (PFA) and Solomonoff-based Fitness Algorithm (SFA) adopt fitness functions based, respectively, on parsimony ideas and on Solomonoff probability induction concepts. These methods are compared in four datasets generated from known functions, with five different levels of additive noise. The experimental analysis indicates that the SFA achieves smaller error when compared to PFA for all the datasets and levels of noise.

Imada and Ross \cite{imada2008using} also present a fitness function, alternative to functions based on the sum of errors, where the scores are determined by the sum of the normalized differences between the target and evaluated values, regarding different statistical features. The experimental analysis in two datasets with two levels of additive noise shows that the proposed fitness function outperforms the fitness based on the sum of errors.


Although the above works handle noise in the symbolic regression context, there is a lack of studies directed to quantify the impact of the noise in GP-based regression methods. The next section presents measures adopted to quantify the influence of noise in classification algorithms from the machine learning literature. In Section \ref{sec:methodology} we select---and adapt---these metrics to apply to our regression test bed.

\subsection{Quantifying Noise Robustness}\label{sec:robustness-mestrics-class}


When a machine learning method is capable of inducing models that are not influenced by the presence of noise in data, we say it is robust to noise---i.e., the more robust a method is to noise, the more similar are the models it induces from data with and without noise \cite{saez2016evaluating}.

Following this premise, works in the classification literature adopt measures that compare the performance of models induced in the presence and absence of noise in the dataset, in order to evaluate the robustness of the learner. Here we introduce three of these metrics: 
relative risk bias, relative loss of accuracy and equalized loss of accuracy.

The Relative Risk Bias (RRB) \cite{kharin1994robustness} measures the robustness of an optimal decision rule---i.e., the Bayesian Decision rule providing the minimal risk when the training data has no ``contaminations''. S\'aez \etal{} \cite{saez2016evaluating} extend the measure to any classifier, given by:

\begin{equation} 
	\label{eq:rrb}
	RRB_{x\%} = \frac{R_{x\%} - R}{R}\enspace,
\end{equation}

\noindent
where $R_{x\%}$ is the classification error rate obtained by the classifier in a dataset with noise level given by $x\%$ and $R$ is the classification error rate of the Bayesian Decision rule without noise (this is a theoretical decision rule, not learned from the data and depends on the data generating process), which is by definition the minimum expected error that can be achieved by any decision rule.

The Relative Loss of Accuracy (RLA) \cite{saez2011fuzzy}, in turn, quantifies the impact of increasing levels of noise in the accuracy of the classifier model when compared to the case with no noise. The RLA measure with level of noise equals to $x\%$ is defined by:

\begin{equation} 
	\label{eq:rla}
	RLA_{x\%} = \frac{A_{0\%} - A_{x\%}}{A_{0\%}}\enspace,
\end{equation}

\noindent
where $A_{0\%}$ and $A_{x\%}$ are the accuracies of the classifier with a noise level of $0\%$ and $x\%$, respectively. RLA is considered more intuitive than RRB, as methods obtaining high values of accuracy without noise ($A_{0\%}$) will have a low RLA value.

Finally, the Equalized Loss of Accuracy (ELA) \cite{saez2016evaluating} was proposed as a correction of the RLA inspired by the measure from
\cite{kharin1994robustness}, and overcomes the limitations of RRB and RLA. The initial performance ($A_{0\%}$) has 
a very low influence in the RLA equation, which can negatively bias the loss of accuracy of methods with high $A_{0\%}$ when compared to 
methods with low initial accuracy. 
E.g., let $A_{0\%}=A_{10\%}=0.5$ be the accuracies of the
method $\alpha$ and $A'_{0\%}=0.8$ and $A'_{10\%}=0.75$ be the accuracies of the method $\beta$. Although method $\beta$ has very low 
loss of accuracy for $10\%$ of noise, the $\alpha$ classifier has a better $RLA_{10\%}$---equals to 0.
The ELA measure is given by:

\begin{equation} 
	\label{eq:ela}
	ELA_{x\%} = \frac{100 - A_{x\%}}{A_{0\%}}\enspace,
\end{equation}

\noindent
where $A_{x\%}$ and $A_{0\%}$ are defined as in Equation \ref{eq:rla}. $ELA_{x\%}$ is equivalent to $RLA_{x\%}+f(A_{0\%})$---see \cite{saez2016evaluating} for the derivation---where the factor $f(A_{0\%})=(100 - A_{0\%})/A_{0\%}$ is equivalent to $ELA_{0\%}$ and depends only on the initial accuracy $A_{0\%}$. Thus the $ELA_{x\%}$ value of a method is based on its robustness, measured by the $RLA_{x\%}$, and on the behavior for clean data---i.e., without controlled noise---measured by $ELA_{0\%}$.

%% file: 4-methodology.tex
\section{Methodology}\label{sec:methodology}

This section presents the methodology followed to analyze how GSGP performs in symbolic regression problems with different levels of noise when compared to GP. We present the datasets considered in our study, along with the strategy to incrementally add noise to the data, and the measures we adopt to assess the impact of different levels of noise on the performance of GSGP and GP.

\subsection{Test Bed}\label{sec:testbed}

Since real-world problems have intrinsic noise inserted when the data is acquired and pre-processed from the environment \cite{nettleton2010study}, we adopt a test bed composed of synthetic data, generated from 15 known functions selected from 
the list of benchmark candidates for symbolic regression GP presented in \cite{McDermott2012benchmarks}.
Table \ref{tab:datasets} presents the function set, the sampling strategy adopted to build the dataset, the input domain, the number of instances and the source from the literature.

\begin{table*}[tp]
	\centering
	\caption{Datasets used in the experiments. 
	Testing and training sets are independent.}
	\scriptsize
	\setlength{\extrarowheight}{2pt}
		\begin{tabular*}{\textwidth}{@{\extracolsep{\fill}}llcABrrl}
			\toprule
			\multicolumn{1}{c}{\multirow{2}{*}{\textbf{Dataset}}} & \multicolumn{1}{c}{\multirow{2}{*}{\textbf{Objective function}}} & \multicolumn{1}{c}{\multirow{2}{*}{\textbf{\# of variables}}} & \multicolumn{2}{c}{\textbf{Sampling strategy}} & \multicolumn{2}{c}{\textbf{\# of instances}} & \multicolumn{1}{c}{\multirow{2}{*}{\textbf{Src.}}}\\
			 &&& \multicolumn{1}{c}{\textbf{Training}} & \multicolumn{1}{c}{\textbf{Test}} & \multicolumn{1}{c}{\textbf{Training}} & \multicolumn{1}{c}{\textbf{Test}}&\\
			 \midrule
			Keijzer-1& $0.3\ x \sin (2\pi x)$ & 1 & $E[-1, 1, 0.1]$ & $E[-1, 1, 0.001]$ & 21 & 2001 & \cite{keijzer2003improving}\\
			Keijzer-2& $0.3\ x \sin (2\pi x)$ & 1 & $E[-2, 2, 0.1]$ & $E[-2, 2, 0.001]$ & 41 & 4001 & \cite{keijzer2003improving}\\
			Keijzer-3& $0.3\ x \sin (2\pi x)$ & 1 & $E[-3, 3, 0.1]$ & $E[-3, 3, 0.001]$ & 61 & 6001 & \cite{keijzer2003improving}\\
			Keijzer-4& $x^3\ e^{-x} \cos (x) \sin(x) (\sin^2 (x) \cos (x) - 1)$ & 1 & E[0, 10, 0.1] & E[0.05, 10.05, 0.1] & 101 & 101 & \cite{keijzer2003improving}\\
			Keijzer-6& $\sum_{i}^{x} \frac{1}{i}$ & 1 & $E[1, 50, 1]$ & $E[1, 120, 1]$ & 50 & 120 & \cite{keijzer2003improving}\\
			Keijzer-7& $ln\ x$ & 1 & $E[1, 100, 1]$ & $E[1, 100, 0.1]$ & 100 & 991 & \cite{keijzer2003improving}\\
			Keijzer-8& $\sqrt{x}$ & 1 & $E[0, 100, 1]$ & $E[0, 100, 0.1]$ & 101 & 1001 & \cite{keijzer2003improving}\\
			Keijzer-9& $\arcsin (x) \quad \quad i.e.,\ ln(x + \sqrt{x^2 + 1})$ & 1 & $E[0, 100, 1]$ & $E[0, 100, 0.1]$ & 100 & 2025 & \cite{keijzer2003improving}\\
			Vladislavleva-1& $\frac{e^{-(x - 1)^2}}{1.2+(y - 2.5)^2}$ & 2 & $U[0.3, 4, 100]$ & $E[-0.2, 4.2, 0.1]$ & 100 & 2025 & \cite{Vladislavleva2009order}\\
			Vladislavleva-2& $e^{-x} x^3(\cos (x) \sin (x))(\cos (x) \sin^2 (x) - 1)$ & 1 & $E[0.05, 10, 0.1]$ & $E[-0.5, 10.5, 0.05]$ & 100 & 221 & \cite{Vladislavleva2009order}\\
			Vladislavleva-3& $e^{-x} x^3(\cos (x) \sin (x))(\cos (x) \sin^2 (x) - 1)(y - 5)$ & 2 & \mbox{$x: E[0.05, 10, 0.1]$} \mbox{$y: E[0.05, 10.05, 2]$} & \mbox{$x: E[-0.5, 10.5, 0.05]$} \mbox{$y: E[-0.5, 10.5, 0.5]$} & 600 & 5083 & \cite{Vladislavleva2009order}\\
			Vladislavleva-4& $\frac{10}{5 + (x - 3)^2 + (y - 3)^2 + (z - 3)^2 + (v - 3)^2 + (w - 3)^2}$& 5 & $U[0.05, 6.05,1024]$ & $U[-0.25, 6.35,5000]$ & 1024 & 5000 & \cite{Vladislavleva2009order}\\
			Vladislavleva-5& $30\frac{(x - 1)(z - 1)}{y^2 (x - 10)}$ & 3 & \mbox{$x: U[0.05, 2,300]$} \mbox{$y: U[1, 2,300]$} \mbox{$z: U[0.05, 2,300]$} & \mbox{$x: E[-0.05, 2.1, 0.15]$} \mbox{$y: E[0.95, 2.05, 0.1]$} \mbox{$z: E[-0.05, 2.1, 0.15]$} & 300 & 2700 & \cite{Vladislavleva2009order}\\
			Vladislavleva-7& $(x - 3)(y - 3) + 2\sin ((x - 4)(y - 4))$ & 2 & $U[0.05, 6.05,300]$ & $U[-0.25, 6.35,1000]$ & 300 & 1000 & \cite{Vladislavleva2009order}\\
			Vladislavleva-8&  $\frac{(x - 3)^4 + (y - 3)^3 - (y - 3)}{(y - 2)^4 + 10}$ & 2 & $U[0.05, 6.05,50]$ & $E[-0.25, 6.35, 0.2]$ & 50 & 1089 & \cite{Vladislavleva2009order}\\
			\bottomrule
		\end{tabular*}
	\label{tab:datasets}
\end{table*}


The training and test sets are sampled independently, according to two strategies presented in Table~\ref{tab:datasets}. $U[a,b, c]$ indicates a uniform random sample of size $c$ drawn from the interval $[a,b]$ and $E[a,b,c]$ indicates a grid of points evenly spaced with an interval $c$, from $a$ to $b$, inclusive.
For the former strategy, we generated five sets of samples 
and for the latter, since the procedure is deterministic, we generated only one sample.

In order to evaluate the impact of noise on GSGP and GP performances, the response variable (desired output) of the training instances was perturbed by an additive Gaussian noise with zero mean and unitary standard deviation, applied with probability given by $r$. We generated datasets with $r$ varying from $0$ to $0.2$ with steps equal to $0.02$, resulting in 11 different levels of noise, in a total of 165 datasets analyzed.

The performance of the methods in the datasets was measured using the Normalized Root Mean Square Error (NRMSE) \cite{keijzer2003improving,defalco2007parsimony}, given by\footnote{The presented NRMSE equation regards the training set. However, the formula is easily extensible to the test set.}:

\begin{equation}\label{eq:nrmse}
	NRMSE=\frac{RMSE\cdot\sqrt{\frac{n}{n-1}}}{\sigma_t}=
	\sqrt{\frac{\sum\limits_{i=1}^n{(y_i-f(x_i))^2}}{\sum\limits_{i=1}^n{(y_i-\bar{t})^2}}}\enspace,
\end{equation}

\noindent
where 
$\bar{t}$ and $\sigma_t$ are, respectively, the mean and standard deviation of the target output vector $t$ and $f$ is the model (function) induced by the regression method. 
NRMSE is equal to 1 when the model performs equivalently to $\bar{t}$ and equal to 0 when the model perfectly fits the data. We used the normalized version of RMSE to be able to compare results from different levels of noise and datasets in a fair way, as described in the next section.

%


\subsection{Noise Robustness in Regression}\label{sec:robustness-mestrics-regres}

The performance of GSGP and GP in the same datasets with different levels of noise is assessed by the robustness measures presented in Section \ref{sec:robustness-mestrics-class}, namely RLA and ELA, adapted to the regression domain. Instead of using the accuracy---a performance measure for classification methods---we adopted the NRMSE.

Notice that the accuracy is defined in $[0\%, 100\%]$---or $[0,1]$---with higher values meaning better accuracy and, consequently, smaller error. Thus, the larger the RLA or ELA measured values, the less robust is the method to the respective noise level. The NRMSE, on the other hand, is defined in $[0,+\infty)$ and higher values mean greater error. 

In this context, we introduce the Relative Increase in Error (RIE) and Equalized Increase in Error (EIE) measures as alternatives to RLA and ELA, respectively, to quantify the noise robustness in the regression domain. RIE and EIE are given by Equations \ref{eq:rie} and \ref{eq:eie}, respectively, where $E_{x\%}$ is the NRMSE obtained by the model in the dataset with $x\%$ of noise, $E_{0\%}$ is the NRMSE obtained by the model in the dataset with no noise, and a plus one term is added to both denominators in order to avoid division by zero. The higher the values of both measures, the more sensitive the model is to the respective noise level.

\begin{equation} 
	\label{eq:rie}
	RIE_{x\%} = \frac{E_{x\%}-E_{0\%}}{1+E_{0\%}}
\end{equation}

\begin{equation} 
	\label{eq:eie}
	EIE_{x\%} = \frac{E_{x\%}}{1+E_{0\%}}
\end{equation}

Similarly to ELA, we can derive EIE according to Equation \ref{eq:eie-deriv}, such that $EIE_{x\%}$ is equal to  $RIE_{x\%}$ plus a term depending only on the model NRMSE with no noise---given by $EIE_{0\%}$.

\begin{equation}
	\label{eq:eie-deriv}
	EIE_{x\%} = \frac{E_{x\%}}{1+E_{0\%}}=\frac{E_{x\%}+E_{0\%}-E_{0\%}}{1+E_{0\%}} =RIE_{x\%}+EIE_{0\%} 
\end{equation}

%% file: 5-experiments.tex
\section{Experimental Analysis}

\begin{table*}[t]
\caption[]{\textit{P}-values obtained by the statistical analysis of the performances of GP and GSGP. The symbol \eq~indicates the null hypothesis was not discarded and the symbol \up (\down) indicates that GSGP is statistically better (worse) than GP with 95\% confidence.}
	\label{tab:results}
	\centering
	\scriptsize
		\begin{tabular*}{\textwidth}{@{\extracolsep{\fill}}lrrrrrrrrrrrrrrrrrrrrrr}
			\toprule
			& \multicolumn{22}{c}{\textbf{Training instances affected by noise (\%)}}\\ 
			\cmidrule(r{0em}){2-23}
			& \multicolumn{2}{c}{\textbf{0}} & \multicolumn{2}{c}{\textbf{2}} & \multicolumn{2}{c}{\textbf{4}} & \multicolumn{2}{c}{\textbf{6}} & \multicolumn{2}{c}{\textbf{8}} & \multicolumn{2}{c}{\textbf{10}} & \multicolumn{2}{c}{\textbf{12}} & \multicolumn{2}{c}{\textbf{14}} & \multicolumn{2}{c}{\textbf{16}} & \multicolumn{2}{c}{\textbf{18}} & \multicolumn{2}{c}{\textbf{20}}\\
			\midrule
			\textbf{NRMSE} & 0.001 & \up & 0.006 & \up & 0.004 & \up & 0.015 & \up & 0.032 & \up & 0.053 & \multicolumn{1}{c}{\eq} & 0.042 & \up & 0.115 & \multicolumn{1}{c}{\eq} & 0.151 & \multicolumn{1}{c}{\eq} & 0.195 & \multicolumn{1}{c}{\eq} & 0.262 & \multicolumn{1}{c}{\eq}\\
			\textbf{RIE} & \multicolumn{2}{c}{---} & 0.003 & \down & 0.002 & \down & 0.001 & \down & 0.000 & \down & 0.000 & \down & 0.000 & \down & 0.000 & \down & 0.000 & \down & 0.000 & \down & 0.000 & \down\\
			\textbf{EIE} & \multicolumn{2}{c}{---} & 0.015 & \up & 0.021 & \up & 0.126 & \multicolumn{1}{c}{\eq} & 0.300 & \multicolumn{1}{c}{\eq} & 0.381 & \multicolumn{1}{c}{\eq} & 0.402 & \multicolumn{1}{c}{\eq} & 0.467 & \multicolumn{1}{c}{\eq} & 0.381 & \multicolumn{1}{c}{\eq} & 0.598 & \multicolumn{1}{c}{\eq} & 0.885 & \multicolumn{1}{c}{\eq}\\
			\bottomrule
		\end{tabular*}
		
\end{table*}

This section presents the experimental analysis of the performance of GSGP in symbolic regression problems with noisy data. We compare the results with a canonical GP \cite{Banzhaf1998genetic}, using the noise robustness measures introduced in Section \ref{sec:robustness-mestrics-regres} and the 15 datasets presented in Table~\ref{tab:datasets} with 11 different noise levels.
Given the non-deterministic nature of GSGP and GP, each experiment was repeated 50 times. As explained in Section \ref{sec:testbed}, we resampled five times the data obtained randomly by the uniform strategy. In datasets with this sampling strategy, the experiments were repeated 10 times for each sample, resulting in a total of 50 repetitions.

Both GP and GSGP were run with a population of 1000 individuals evolved for 2000 generations with tournament selection of size 10. The grow method \cite{Koza1992genetic} was adopted to generate the random functions inside the geometric semantic crossover and mutation operators, and the ramped half-and-half method \cite{Koza1992genetic} to generate the initial population, both with maximum individual depth equals to 6. The function set included three binary arithmetic operators ($+, -, \times$) and the analytic quotient (AQ) \cite{Ni2013use}, an alternative to the arithmetic division with similar properties, but without discontinuity, given by:

\begin{equation}
	\label{eq:aq}
	\textit{AQ}(a,b)=\frac{a}{\sqrt{1+b^2}}\enspace.
\end{equation}


The terminal set comprised the variables of the problem and constant values uniformly picked from $[-1,1]$. The GP method employed the canonical crossover and mutation operators \cite{Koza1992genetic} with probabilities $0.9$ and $0.1$, respectively.  GSGP employed the geometric semantic crossover for fitness function based on Manhattan distance and mutation operators, as presented in \cite{Castelli2014cpp}, both with probability $0.5$. The mutation step adopted by the geometric semantic mutation operator was defined as 10\% of the standard deviation of the target vector $t$ given by the training data.

\begin{figure*}
	\captionsetup[sub]{skip=1mm}
	\centering
	\begin{subfigure}[t]{0.32\textwidth}
		\centering
		\includegraphics[scale=0.2]{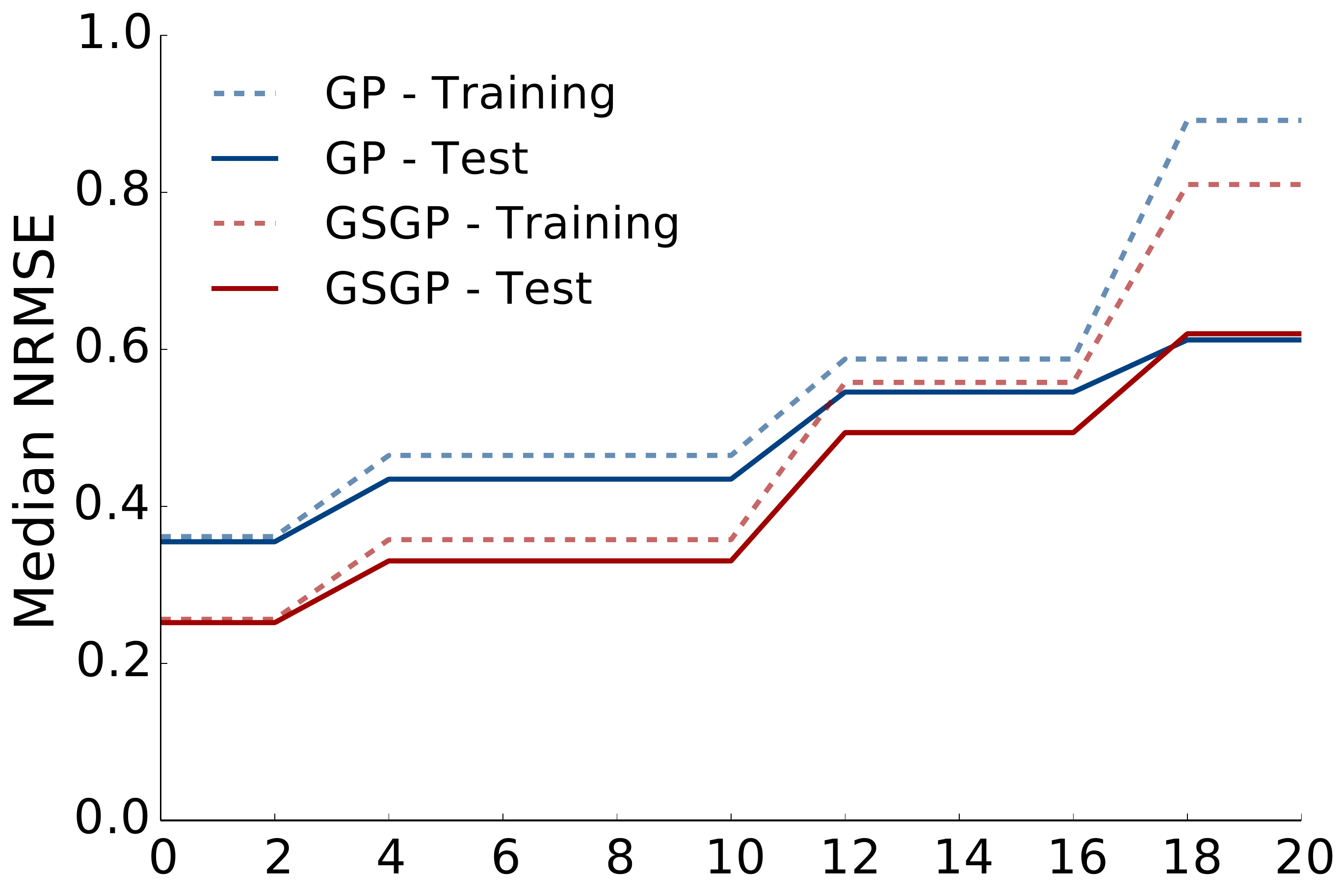}
		\caption{Keijzer-1}\label{fig:nrmse-kei1}
	\end{subfigure}
	\hfill
	\begin{subfigure}[t]{0.3\textwidth}
		\centering
		\includegraphics[scale=0.2]{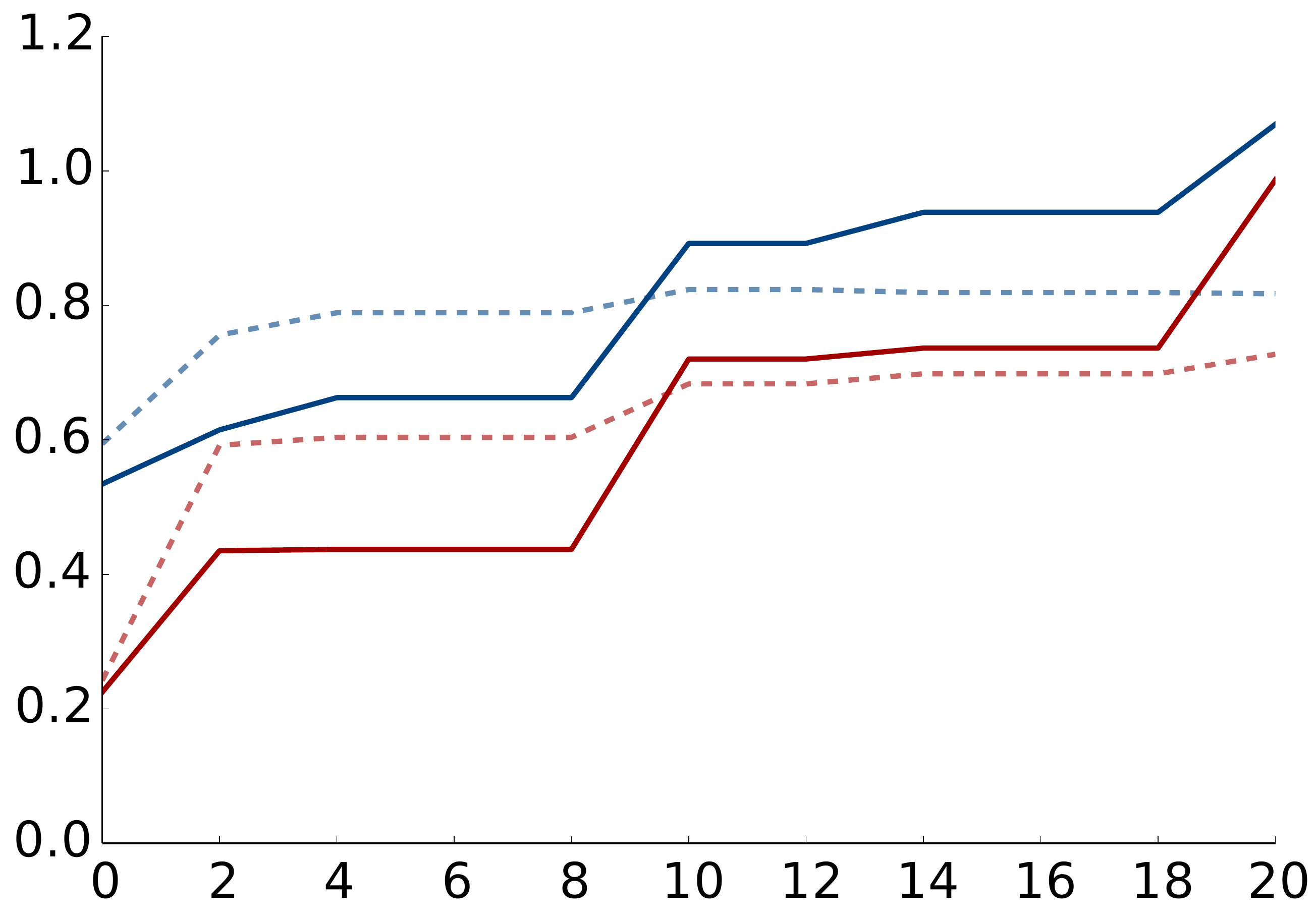}
		\caption{Keijzer-2}\label{fig:nrmse-kei2}
	\end{subfigure}
	\hfill
	\begin{subfigure}[t]{0.3\textwidth}
		\centering
		\includegraphics[scale=0.2]{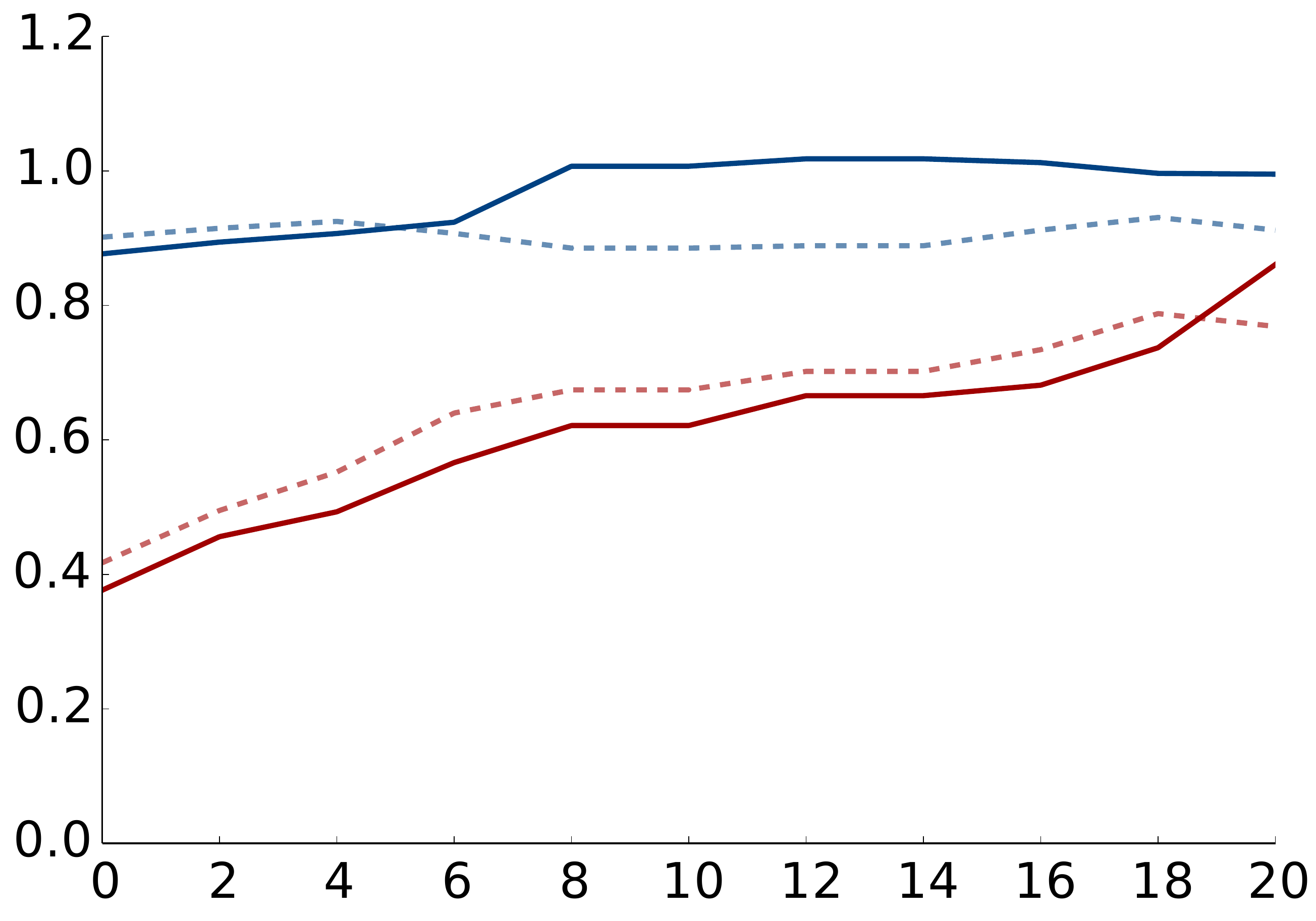}
		\caption{Keijzer-3}\label{fig:nrmse-kei3}
	\end{subfigure}
	\\[0mm]
	\begin{subfigure}[t]{0.32\textwidth}
		\centering
		\includegraphics[scale=0.2]{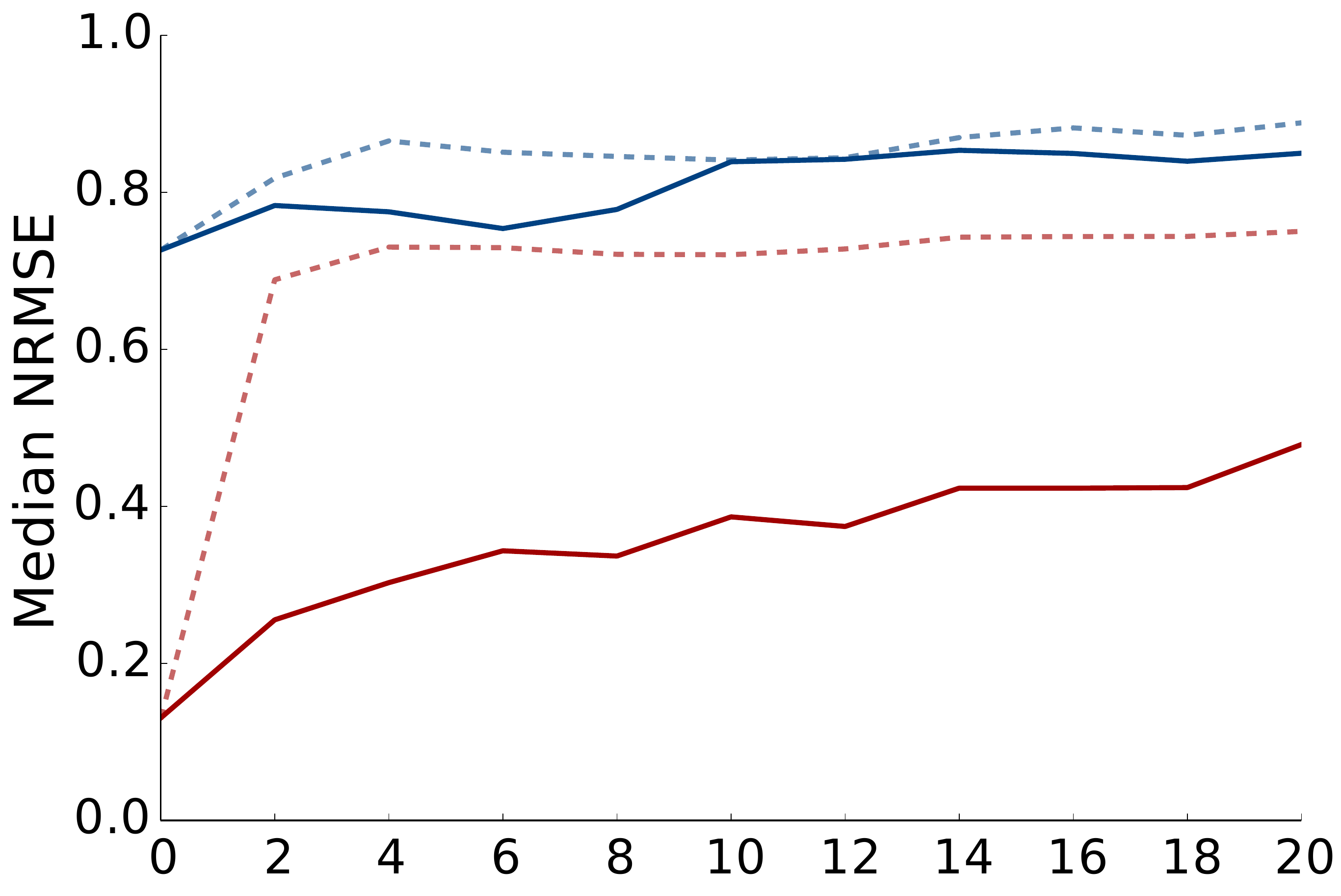}
		\caption{Keijzer-4}\label{fig:nrmse-kei4}
	\end{subfigure}
	\hfill
	\begin{subfigure}[t]{0.3\textwidth}
		\centering
		\includegraphics[scale=0.2]{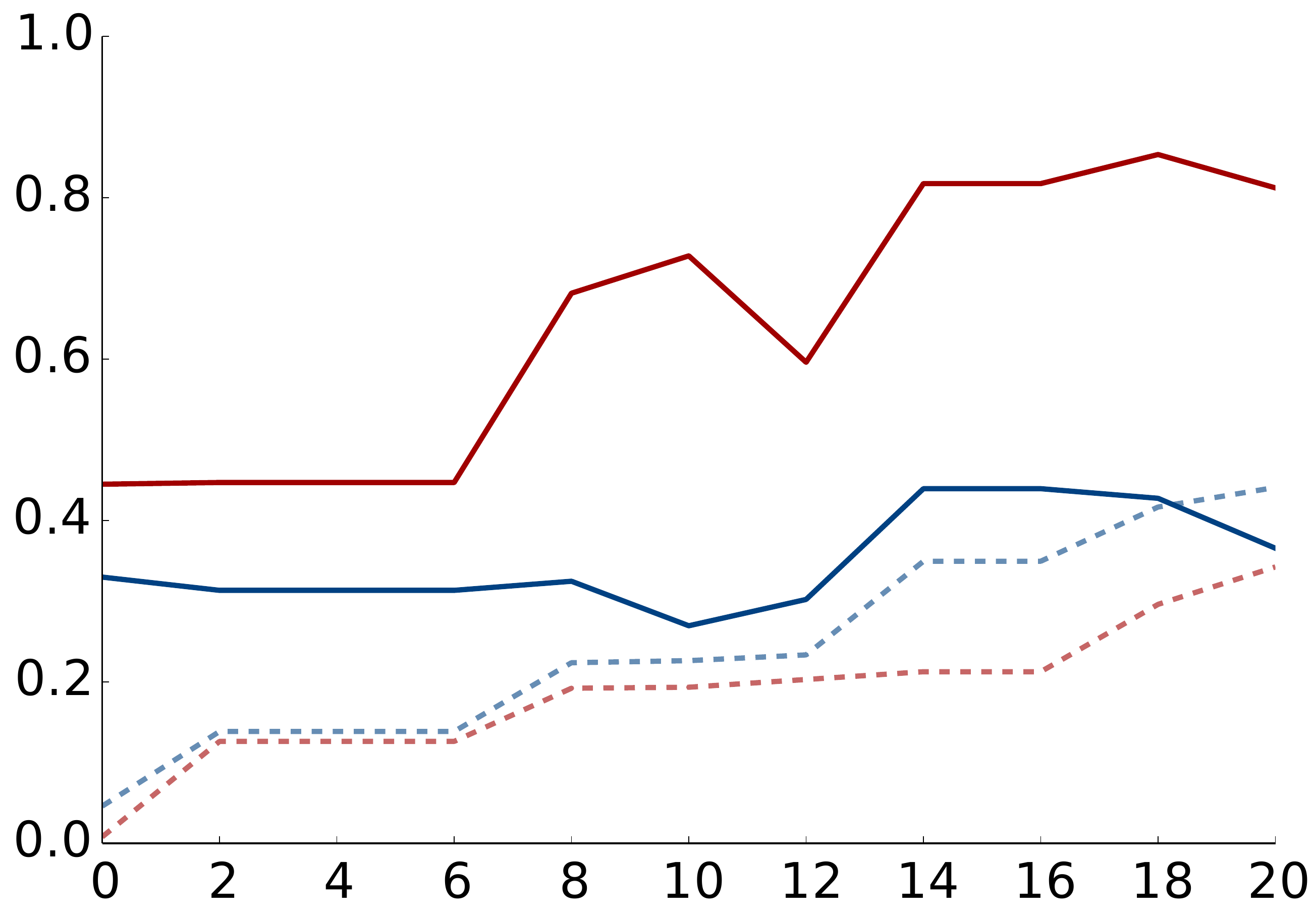}
		\caption{Keijzer-6}\label{fig:nrmse-kei6}
	\end{subfigure}
	\hfill
	\begin{subfigure}[t]{0.3\textwidth}
		\centering
		\includegraphics[scale=0.2]{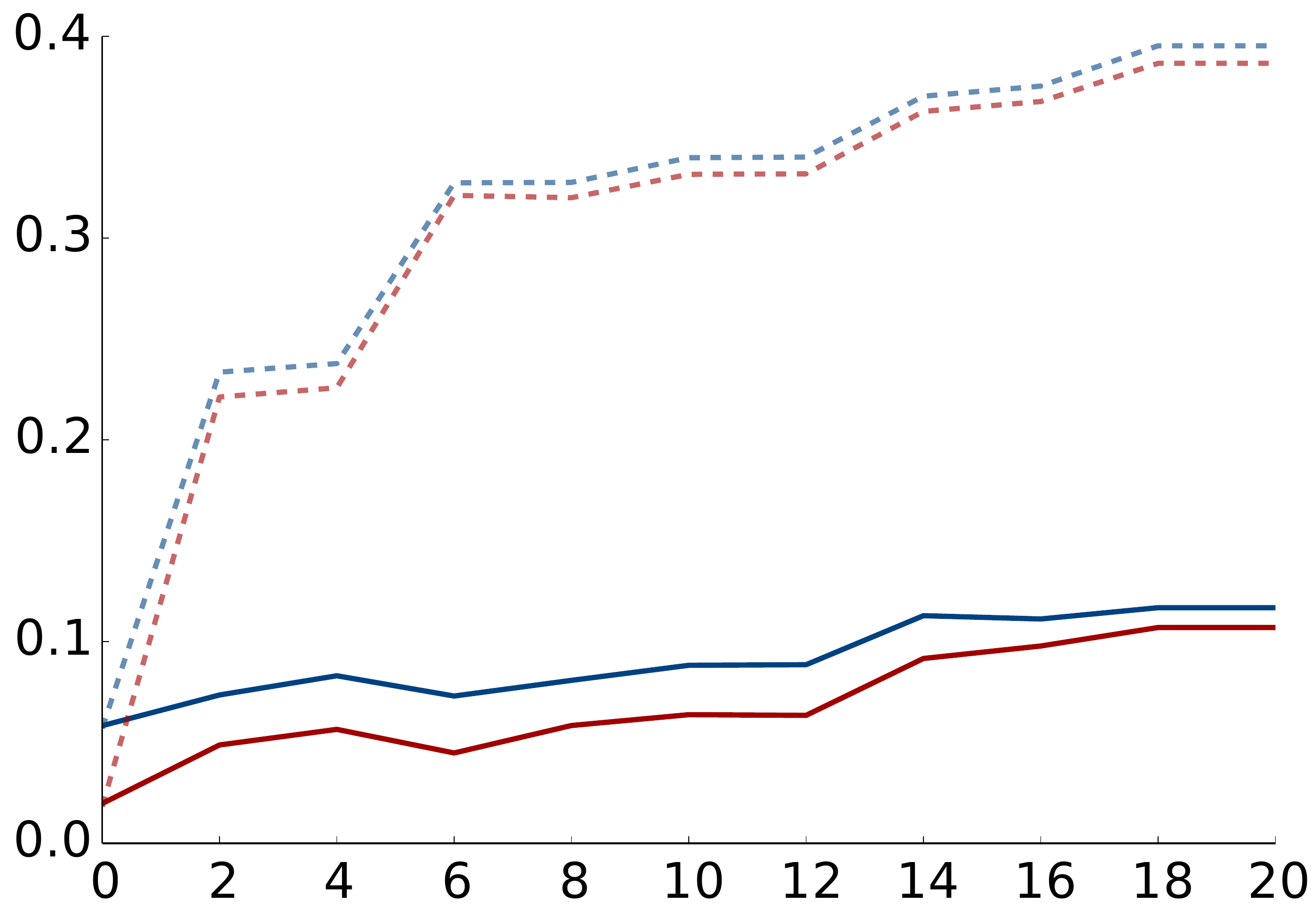}
		\caption{Keijzer-7}\label{fig:nrmse-kei7}
	\end{subfigure}
	\\[0mm]
	\begin{subfigure}[t]{0.32\textwidth}
		\centering
		\includegraphics[scale=0.2]{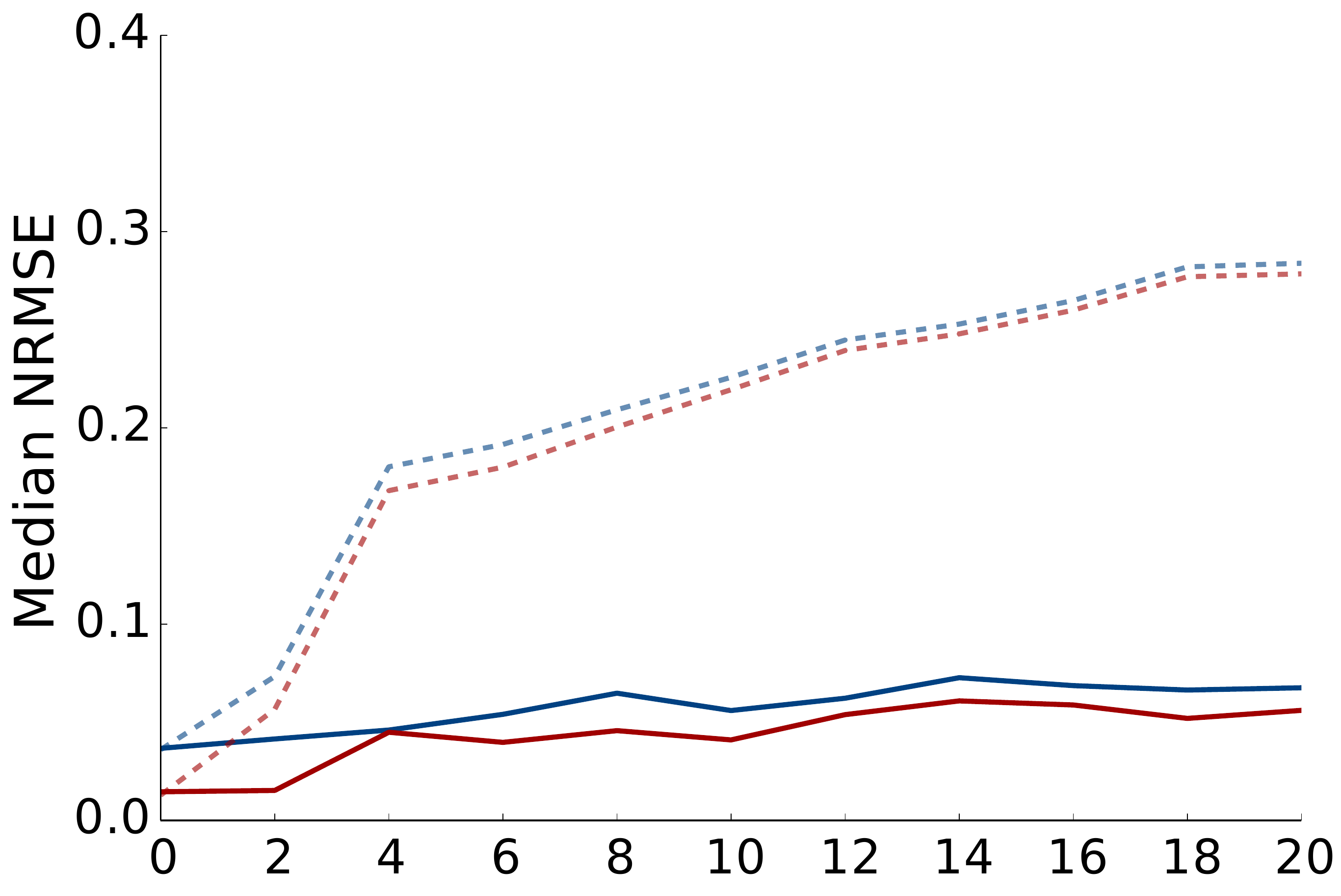}
		\caption{Keijzer-8}\label{fig:nrmse-kei8}
	\end{subfigure}
	\hfill
	\begin{subfigure}[t]{0.3\textwidth}
		\centering
		\includegraphics[scale=0.2]{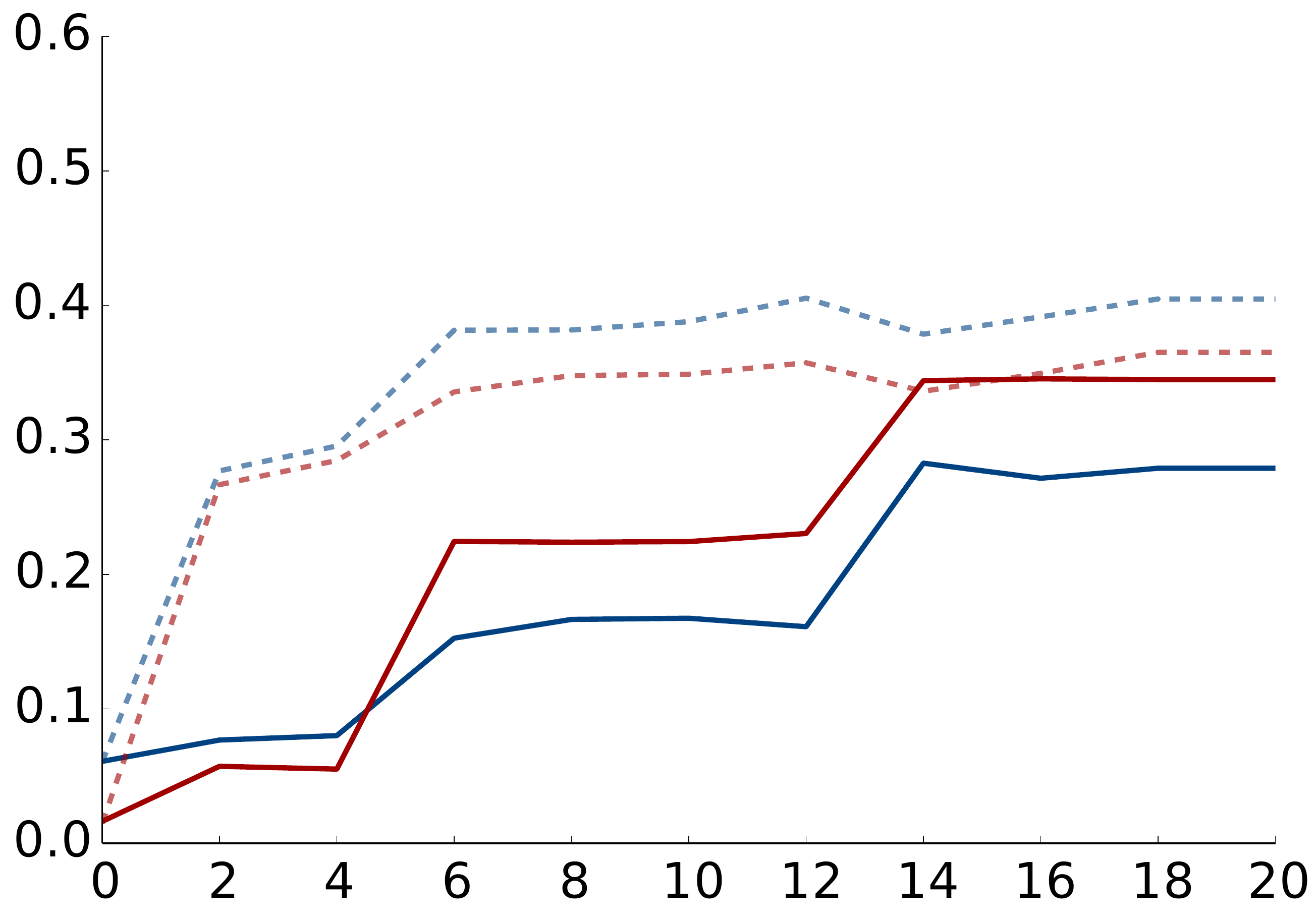}
		\caption{Keijzer-9}\label{fig:nrmse-kei9}
	\end{subfigure}
	\hfill
	\begin{subfigure}[t]{0.3\textwidth}
		\centering
		\includegraphics[scale=0.2]{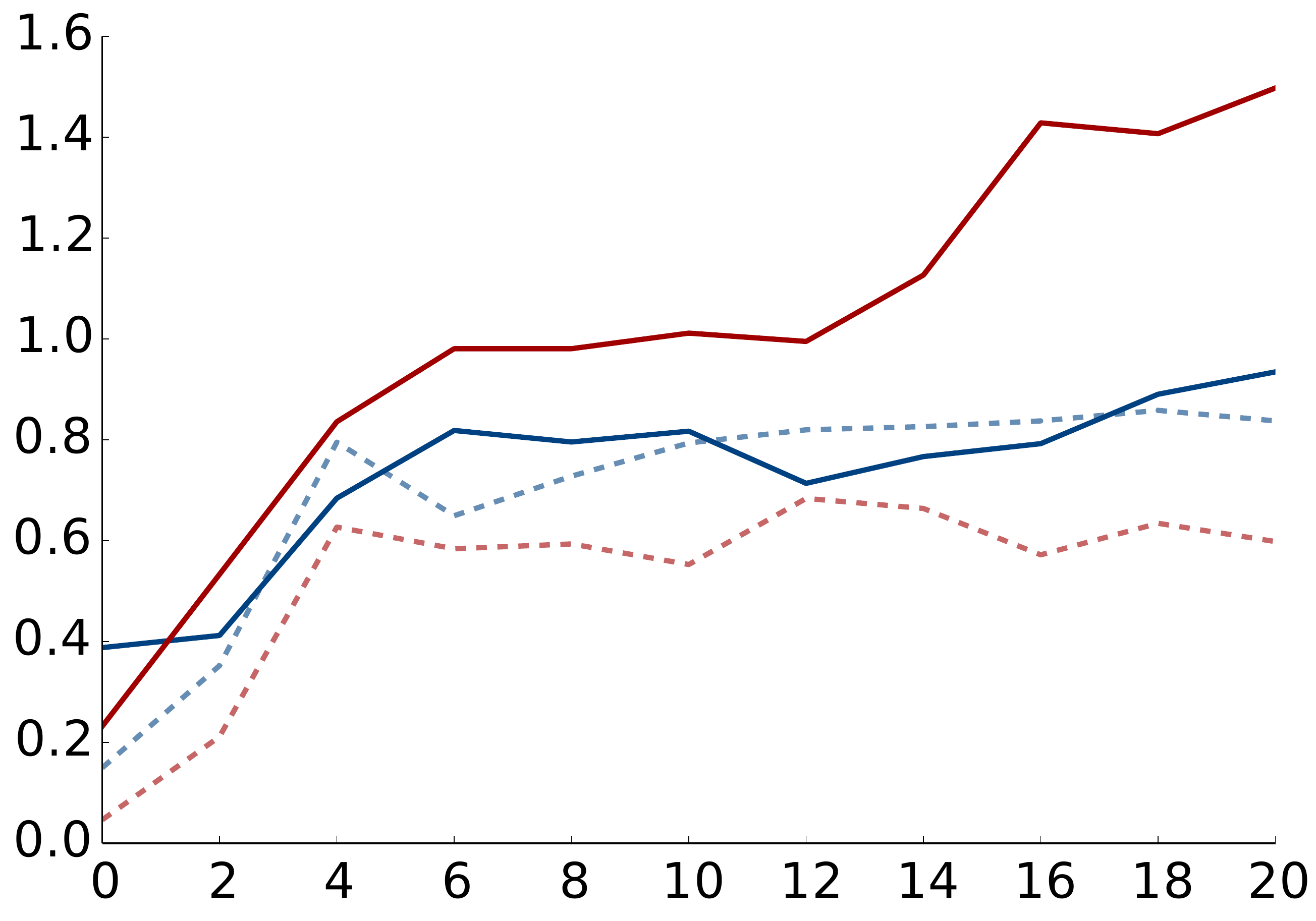}
		\caption{Vladislavleva-1}\label{fig:nrmse-vla1}
	\end{subfigure}
	\\[0mm]
	\begin{subfigure}[t]{0.32\textwidth}
		\centering
		\includegraphics[scale=0.2]{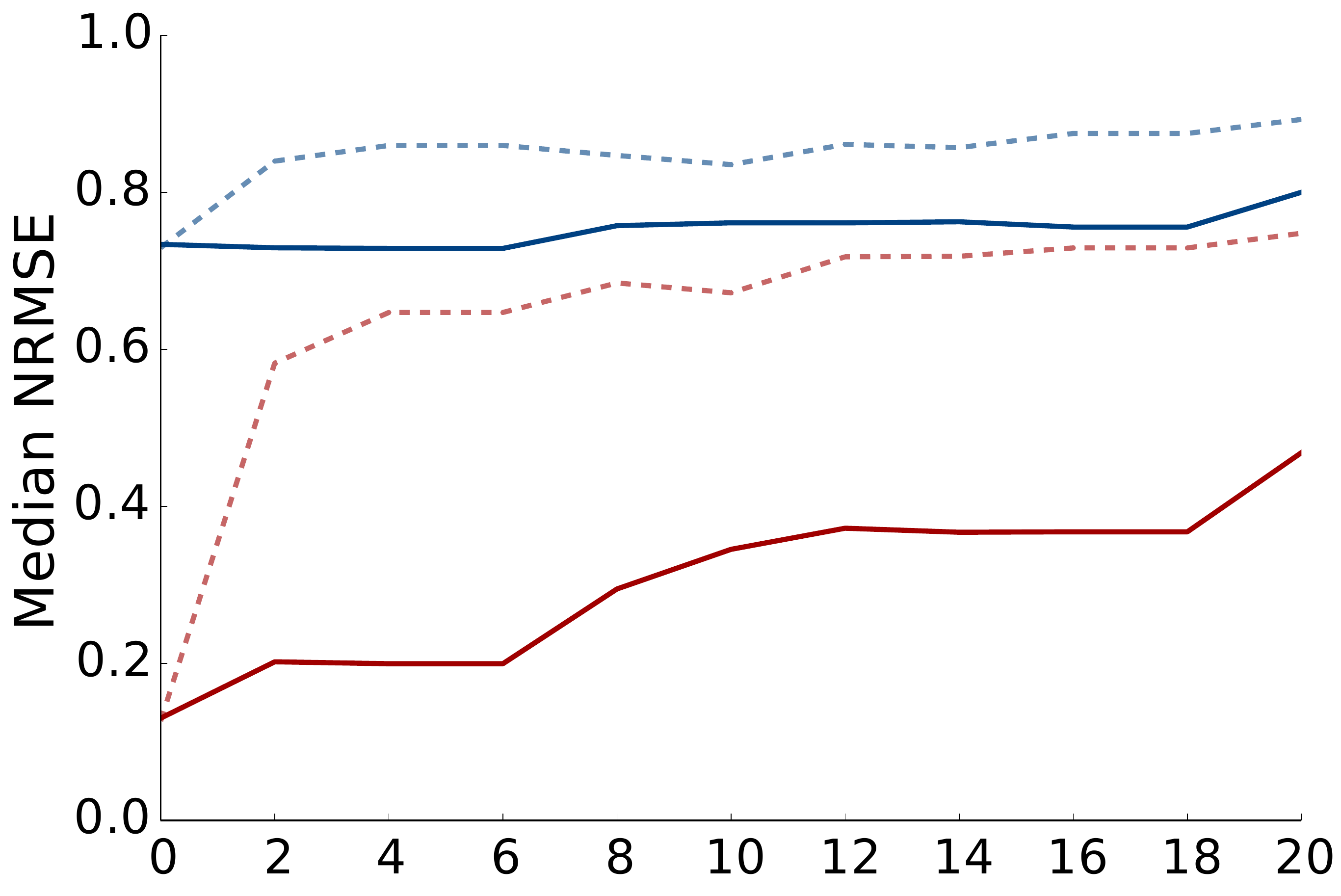}
		\caption{Vladislavleva-2}\label{fig:nrmse-vla2}
	\end{subfigure}
	\hfill
	\begin{subfigure}[t]{0.3\textwidth}
		\centering
		\includegraphics[scale=0.2]{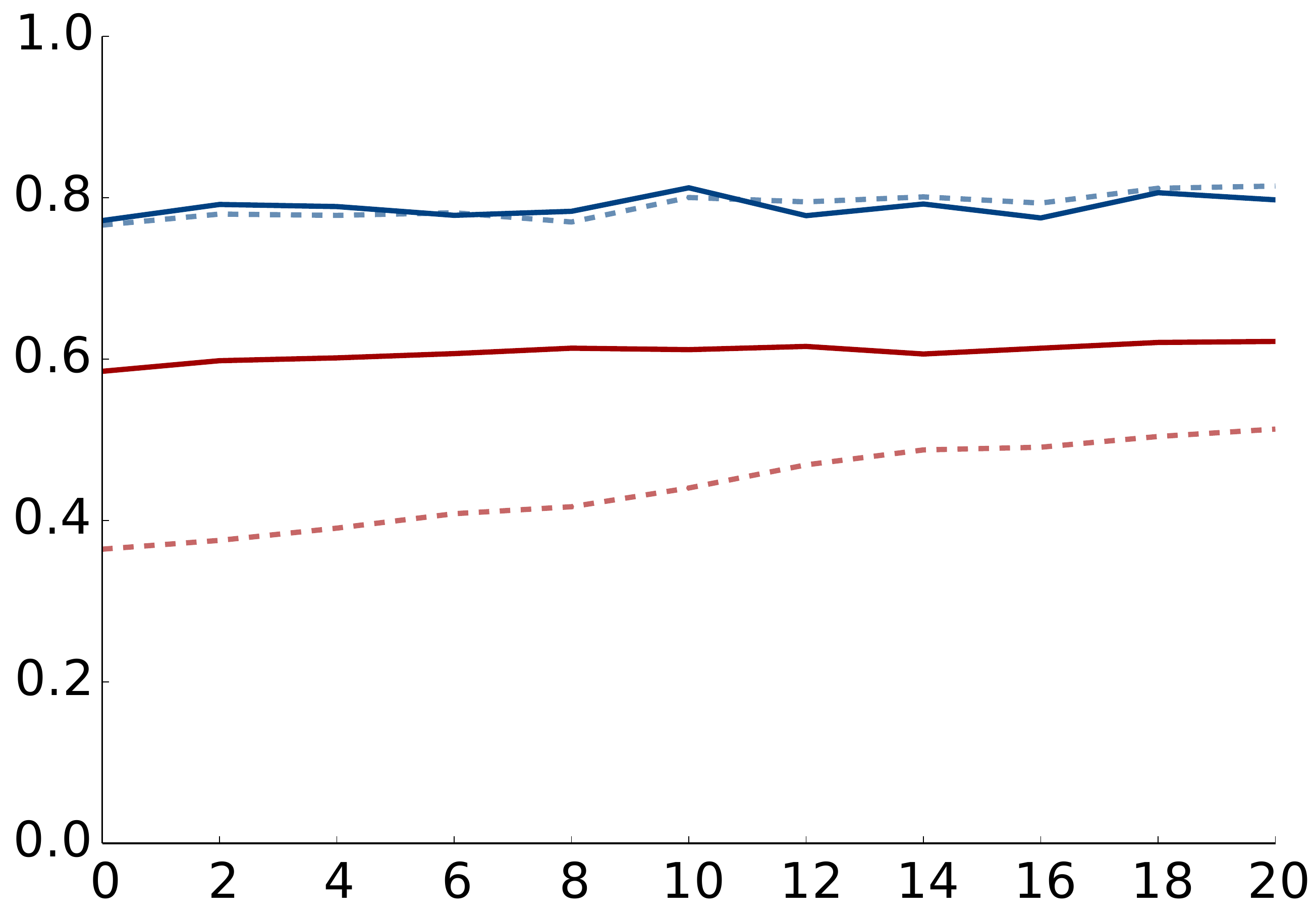}
		\caption{Vladislavleva-3}\label{fig:nrmse-vla3}
	\end{subfigure}
	\hfill
	\begin{subfigure}[t]{0.3\textwidth}
		\centering
		\includegraphics[scale=0.2]{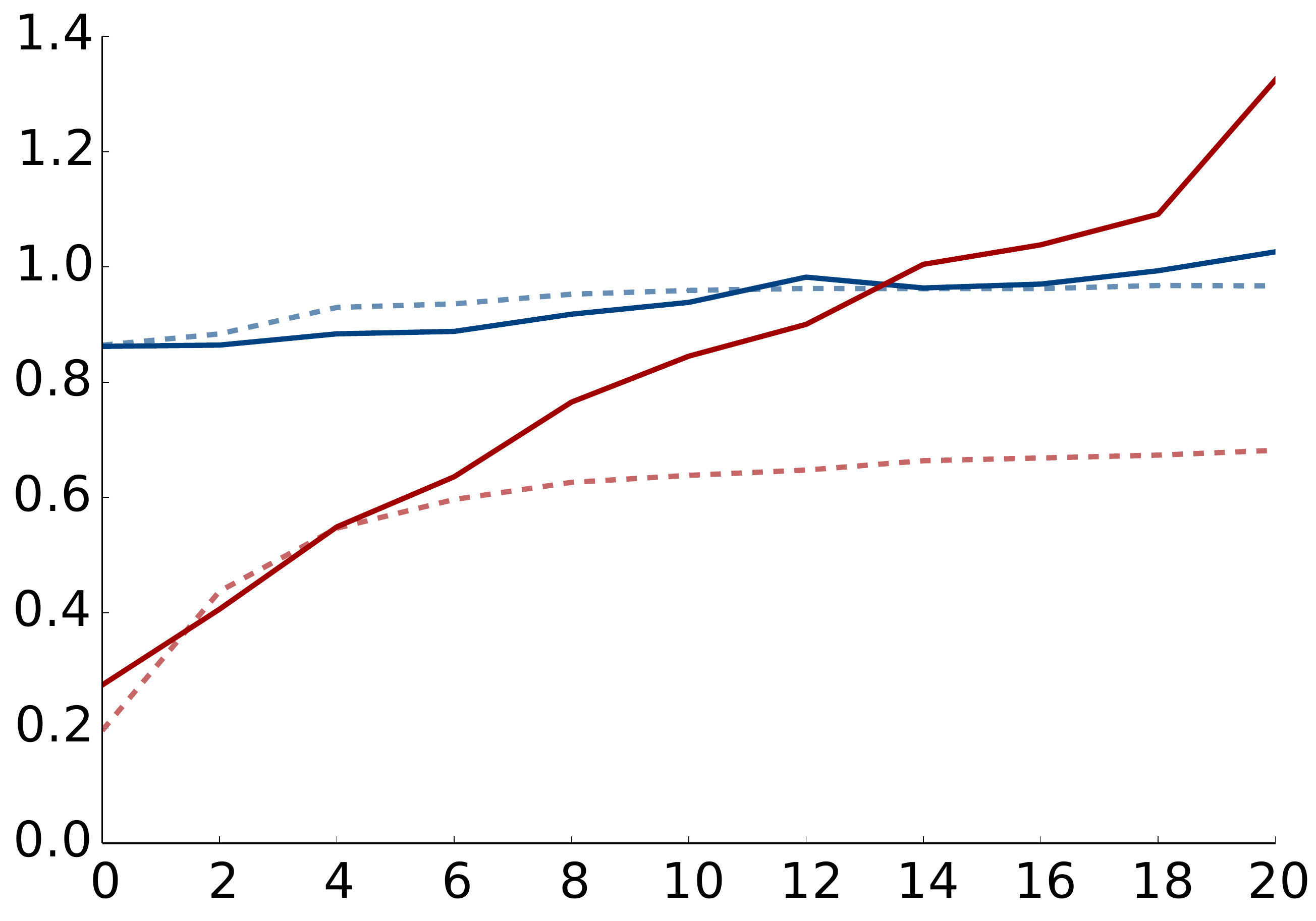}
		\caption{Vladislavleva-4}\label{fig:nrmse-vla4}
	\end{subfigure}
	\\[0mm]
	\begin{subfigure}[t]{0.32\textwidth}
		\centering
		\includegraphics[scale=0.2]{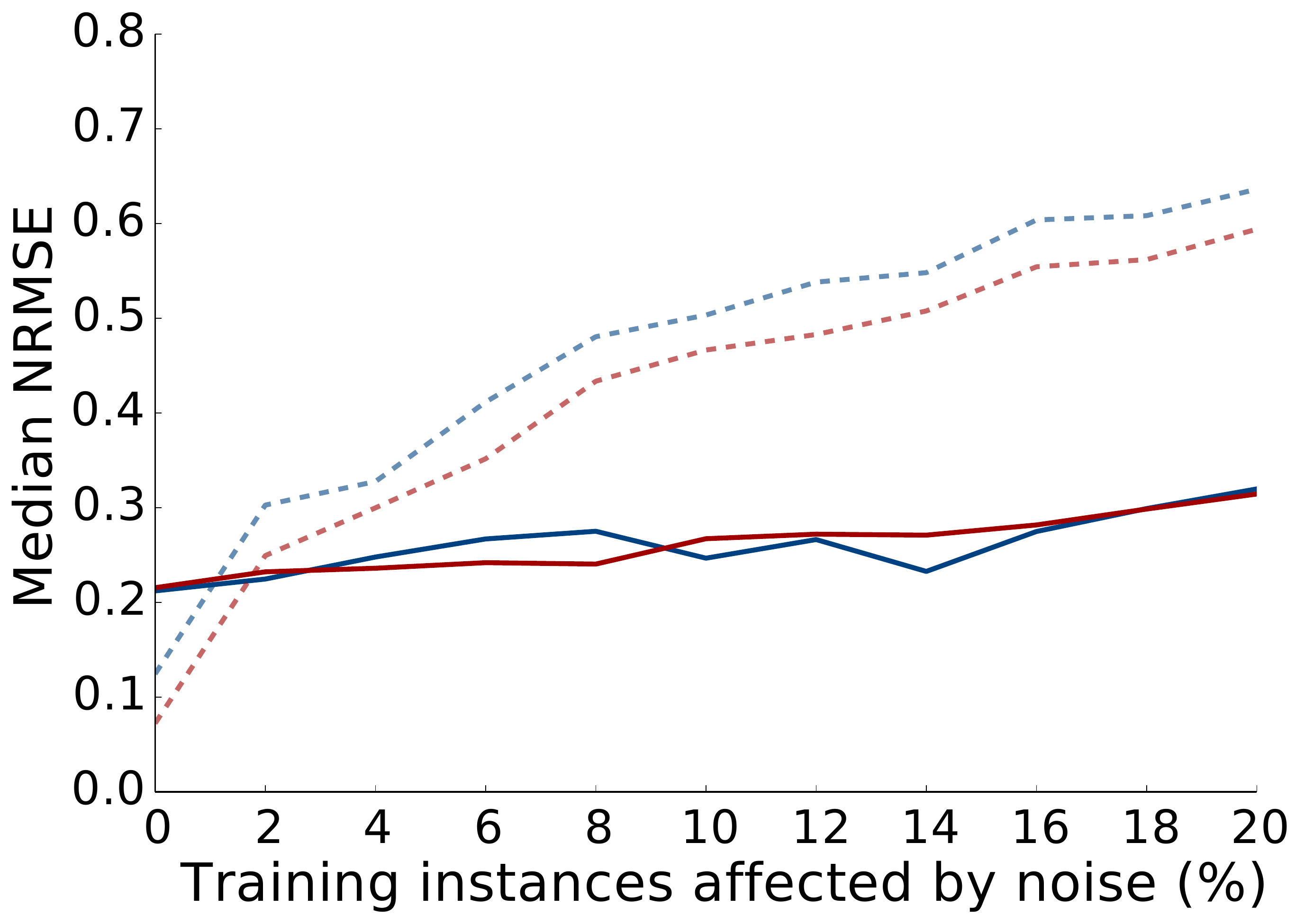}
		\caption{Vladislavleva-5}\label{fig:nrmse-vla5}
	\end{subfigure}
	\hfill
	\begin{subfigure}[t]{0.3\textwidth}
		\centering
		\includegraphics[scale=0.2]{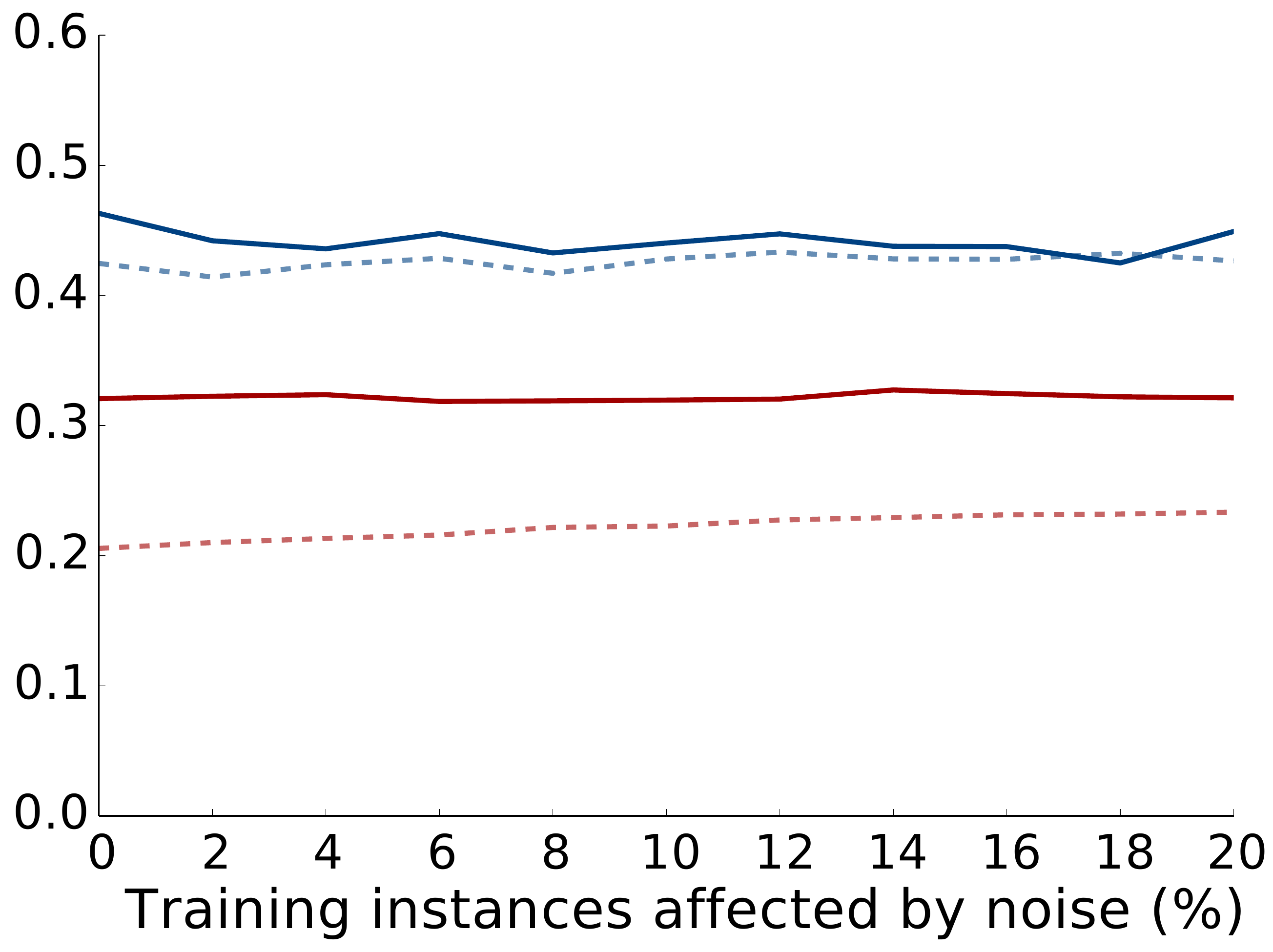}
		\caption{Vladislavleva-7}\label{fig:nrmse-vla7}
	\end{subfigure}
	\hfill
	\begin{subfigure}[t]{0.3\textwidth}
		\centering
		\includegraphics[scale=0.2]{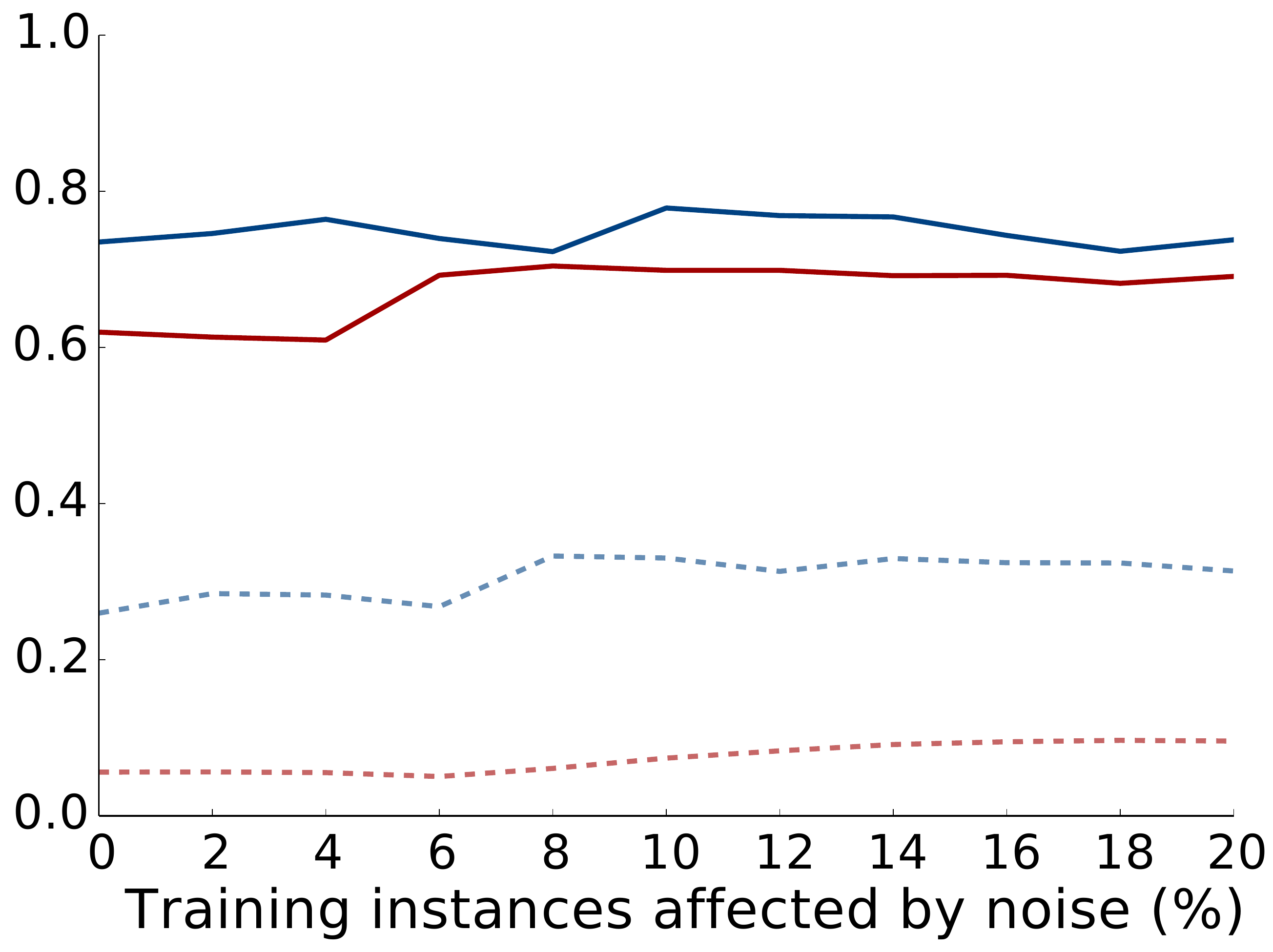}
		\caption{Vladislavleva-8}\label{fig:nrmse-vla8}
	\end{subfigure}
	\caption{Median training and test NRSME obtained by GP and GSGP for each dataset.} 
	\label{fig:nrmse}
\end{figure*}

Figure \ref{fig:nrmse} shows how the median training and test NRMSE are affected when increasing the percentage of noisy instances. Regarding the results for data with no noise, GSGP presents better median test NRMSE in all but two datasets, Keijzer-6 and Vladislavleva-5. However, the opposite behavior is observed for noise levels greater than or equal to $18\%$ in Keijzer-1, $6\%$ in Keijzer-9, $2\%$ in Vladislavleva-1 and $14\%$ in Vladislavleva-4. Moreover, GSGP test NRMSE approximates from GP when the noise level increases in the datasets Keijzer-2, Keijzer-3, Keijzer-4, Keijzer-7, Keijzer-8, Vladislavleva-2 and Vladislavleva-8. This behavior may indicate that, although GSGP outperforms GP in low levels of noise in most of the datasets, its performance deteriorates faster than GP when the level of noise increases. Notice that in all experiments the median training NRMSE of the GSGP is smaller than the one obtained by GP, regardless of the behavior of both methods in the test data, which may indicate that GSGP has a greater tendency to overfit noisy data than GP.

\begin{figure*}
	\captionsetup[sub]{skip=1mm}
	\centering
	\begin{subfigure}[t]{0.32\textwidth}
		\centering
		\includegraphics[scale=0.2]{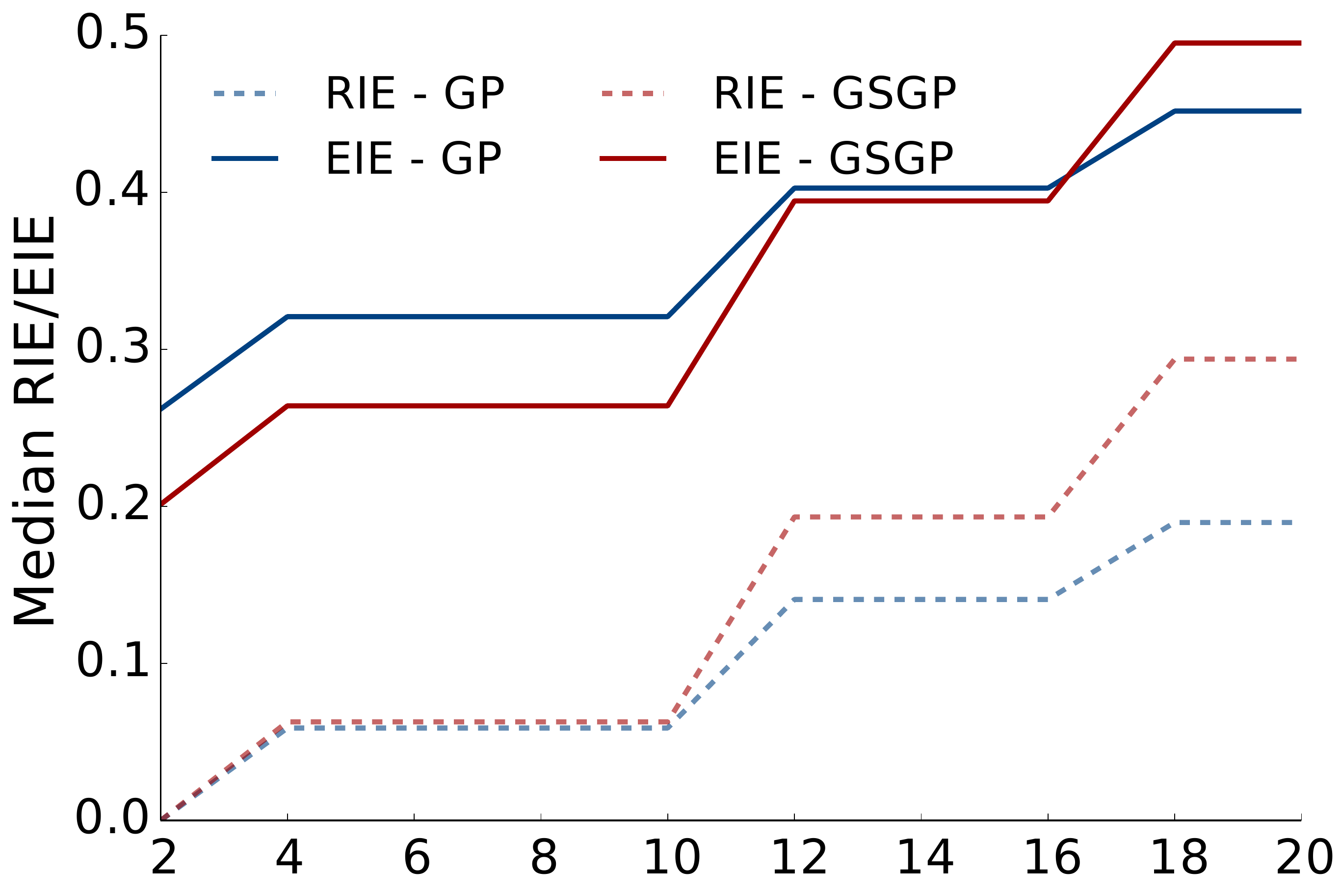}
		\caption{Keijzer-1}\label{fig:rie-eie-kei1}
	\end{subfigure}
	\hfill
	\begin{subfigure}[t]{0.3\textwidth}
		\centering
		\includegraphics[scale=0.2]{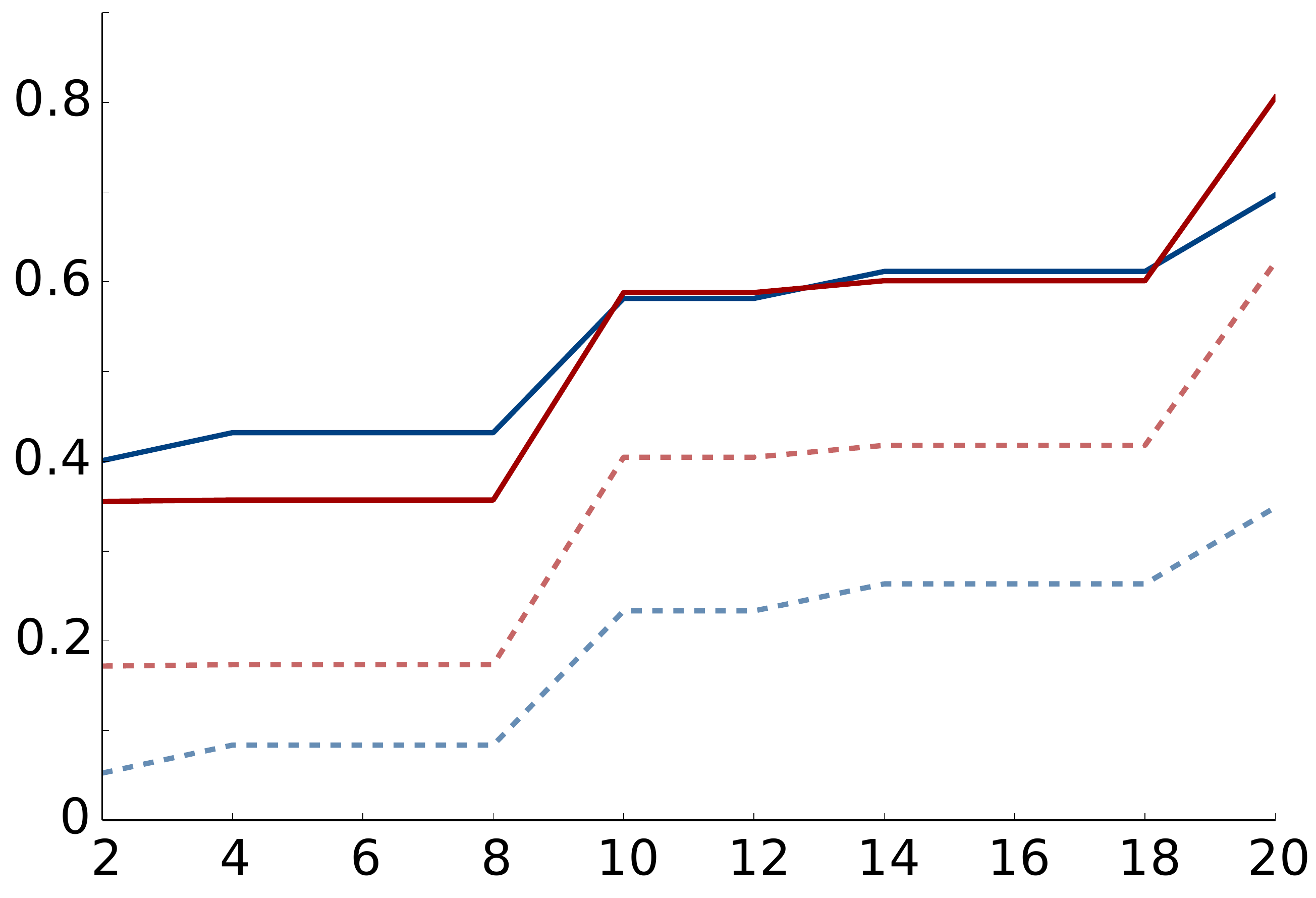}
		\caption{Keijzer-2}\label{fig:rie-eie-kei2}
	\end{subfigure}
	\hfill
	\begin{subfigure}[t]{0.3\textwidth}
		\centering
		\includegraphics[scale=0.2]{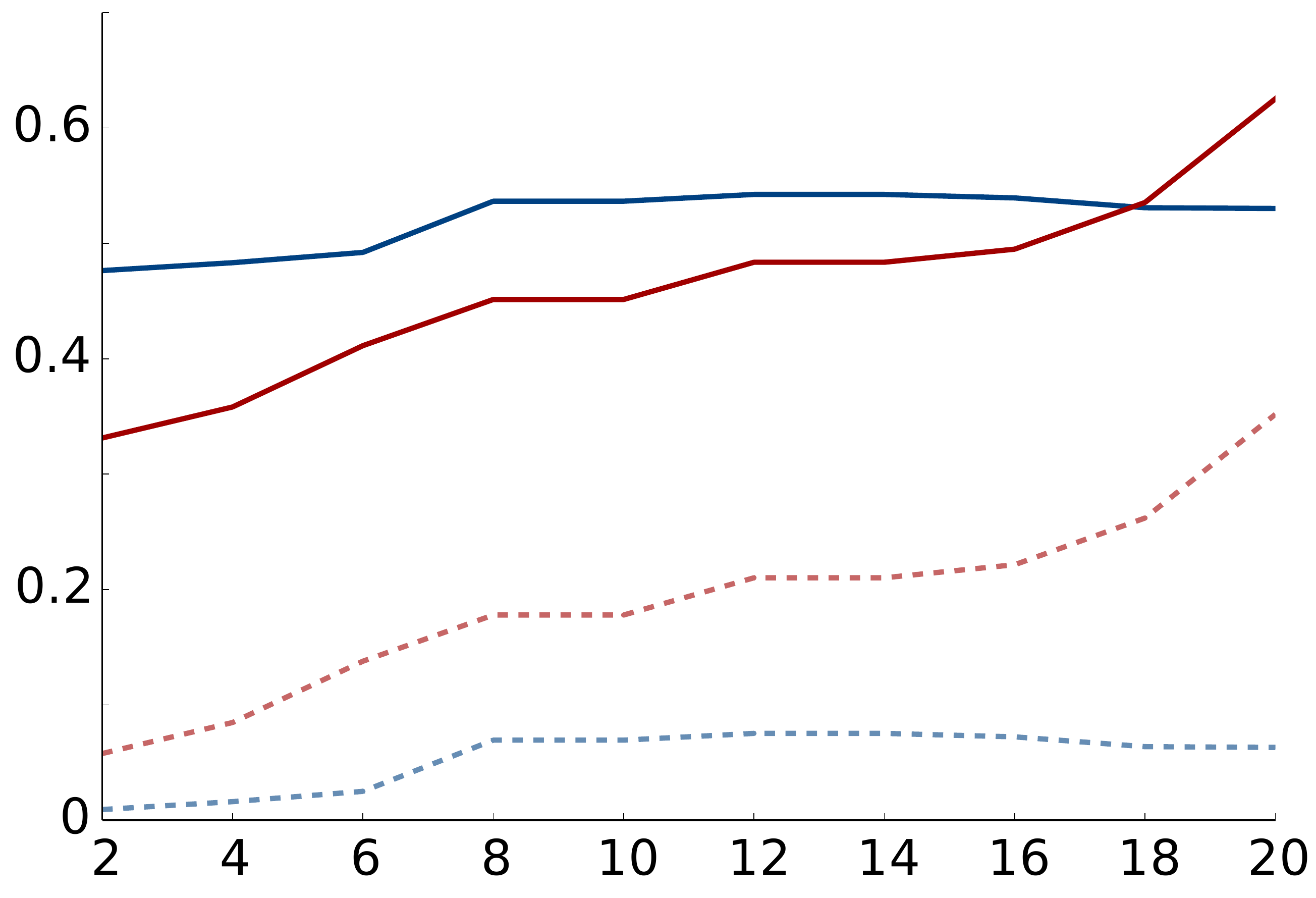}
		\caption{Keijzer-3}\label{fig:rie-eie-kei3}
	\end{subfigure}
	\\[0mm]
	\begin{subfigure}[t]{0.32\textwidth}
		\centering
		\includegraphics[scale=0.2]{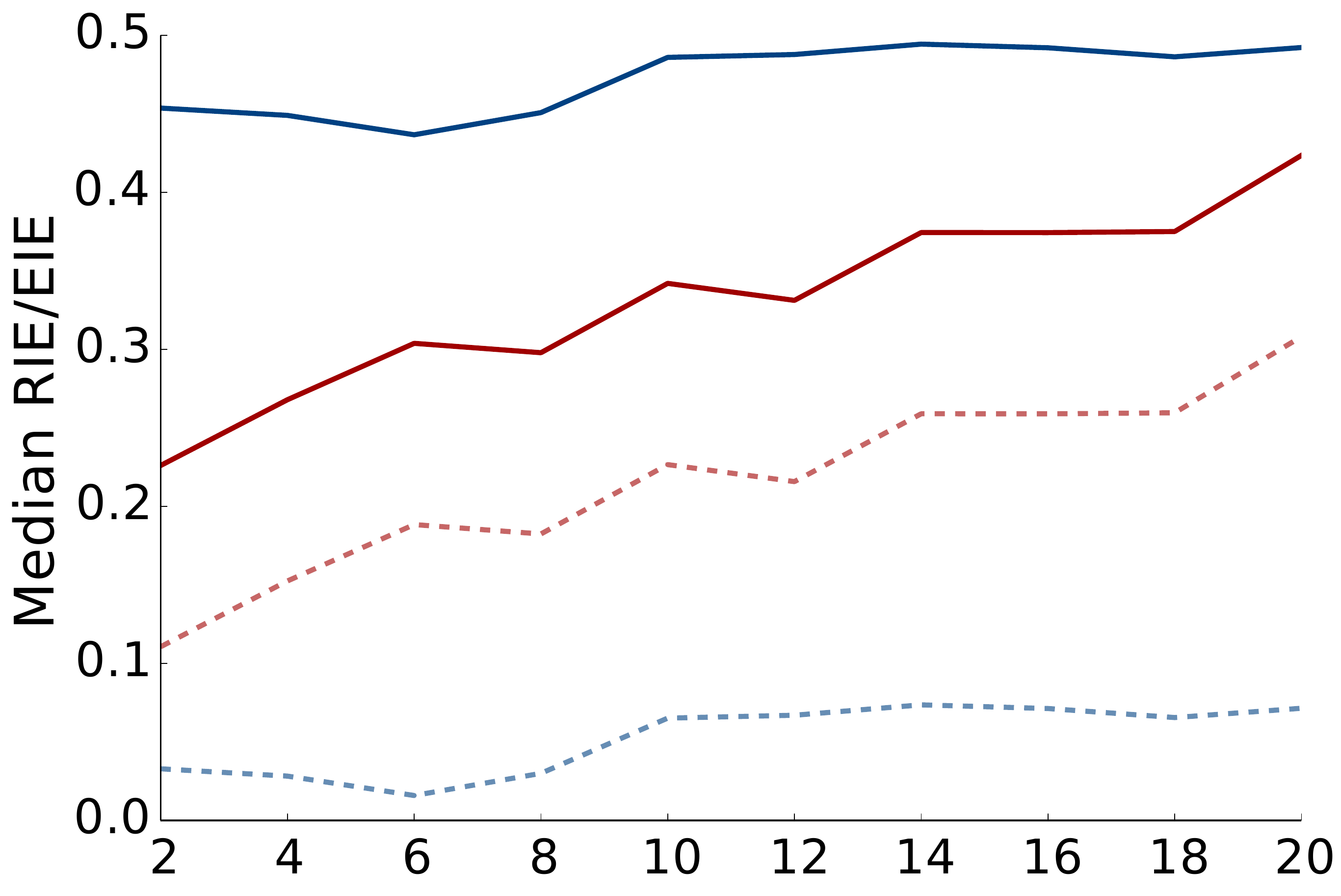}
		\caption{Keijzer-4}\label{fig:rie-eie-kei4}
	\end{subfigure}
	\hfill
	\begin{subfigure}[t]{0.3\textwidth}
		\centering
		\includegraphics[scale=0.2]{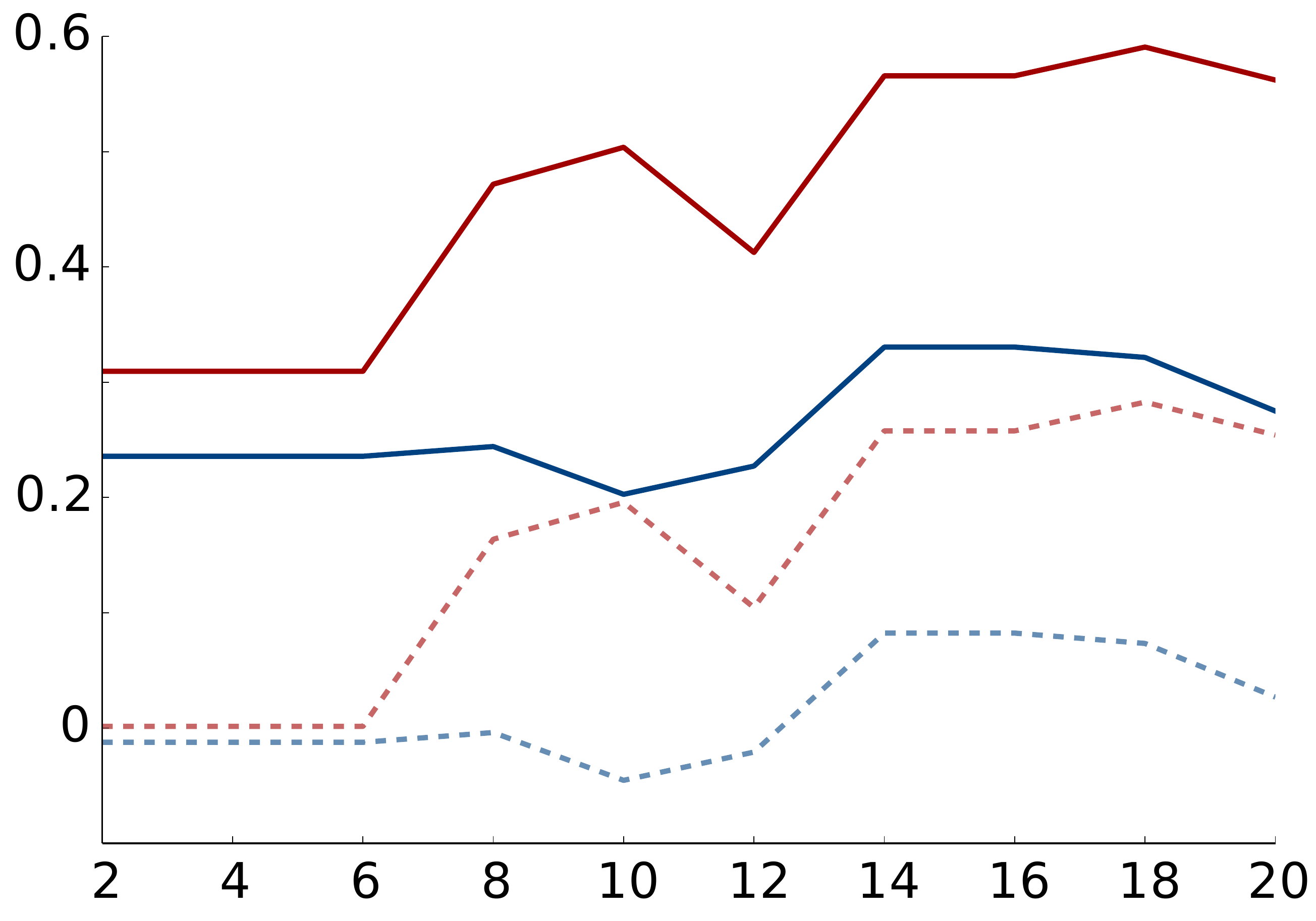}
		\caption{Keijzer-6}\label{fig:rie-eie-kei6}
	\end{subfigure}
	\hfill
	\begin{subfigure}[t]{0.3\textwidth}
		\centering
		\includegraphics[scale=0.2]{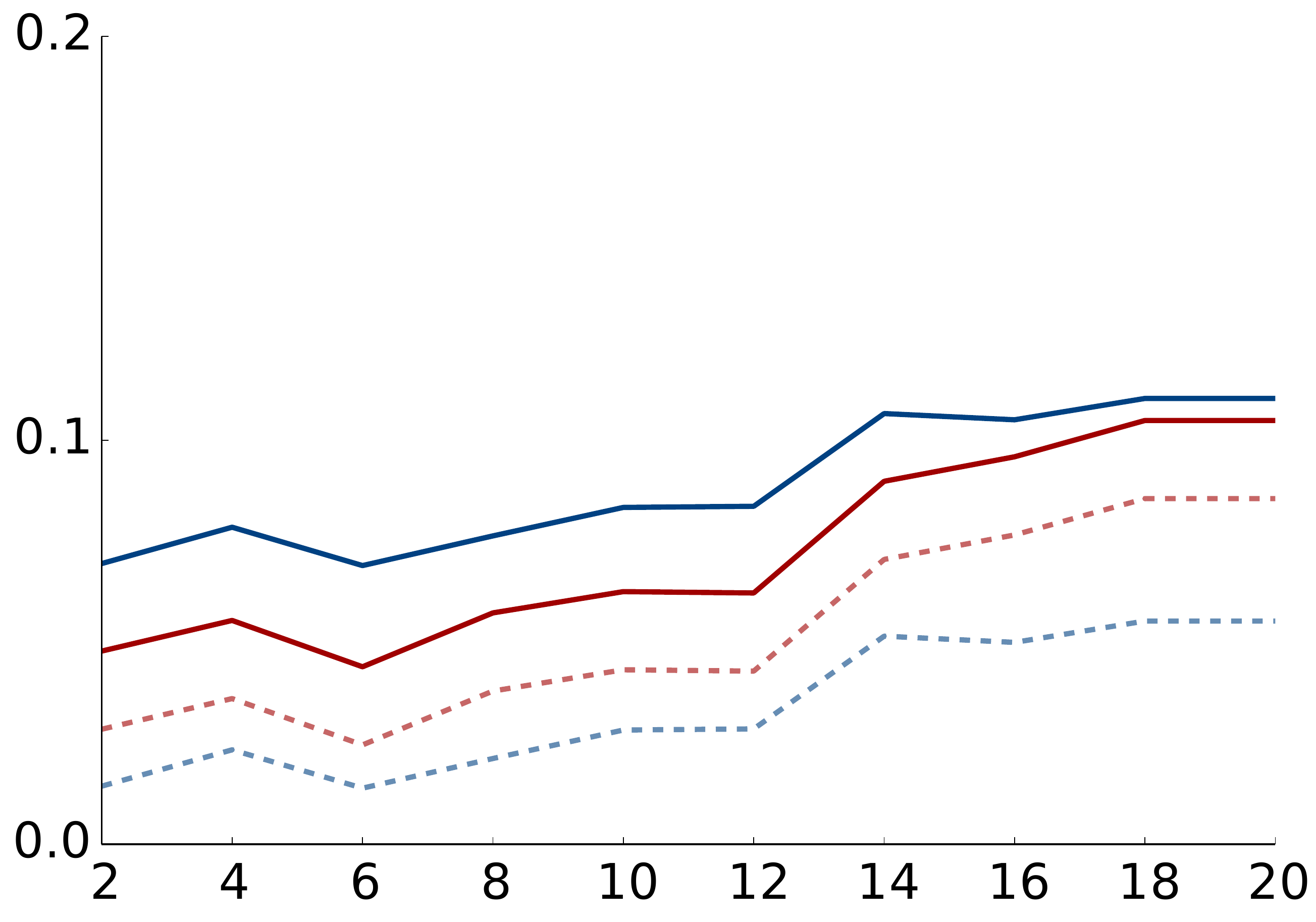}
		\caption{Keijzer-7}\label{fig:rie-eie-kei7}
	\end{subfigure}
	\\[0mm]
	\begin{subfigure}[t]{0.32\textwidth}
		\centering
		\includegraphics[scale=0.2]{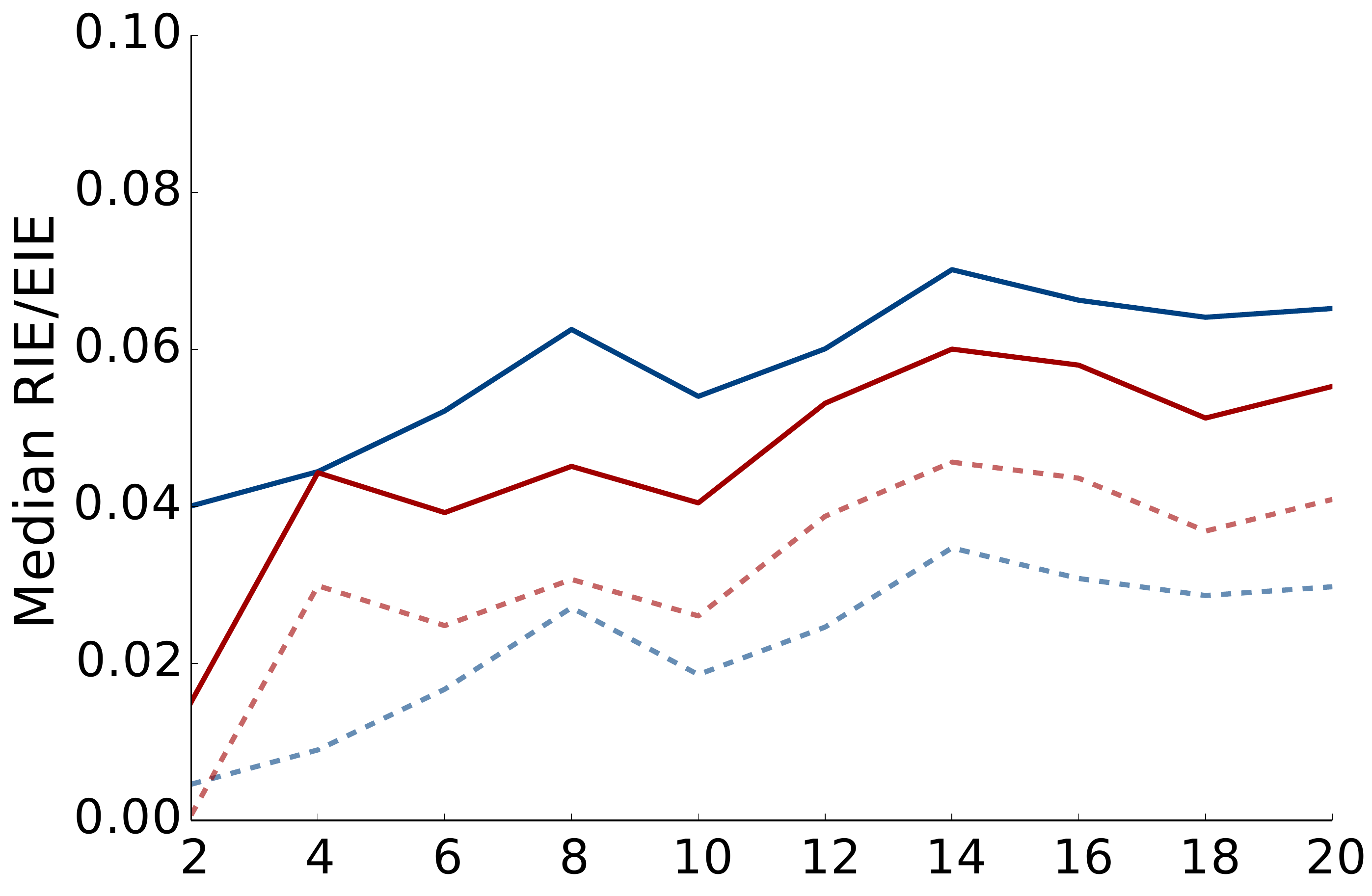}
		\caption{Keijzer-8}\label{fig:rie-eie-kei8}
	\end{subfigure}
	\hfill
	\begin{subfigure}[t]{0.3\textwidth}
		\centering
		\includegraphics[scale=0.2]{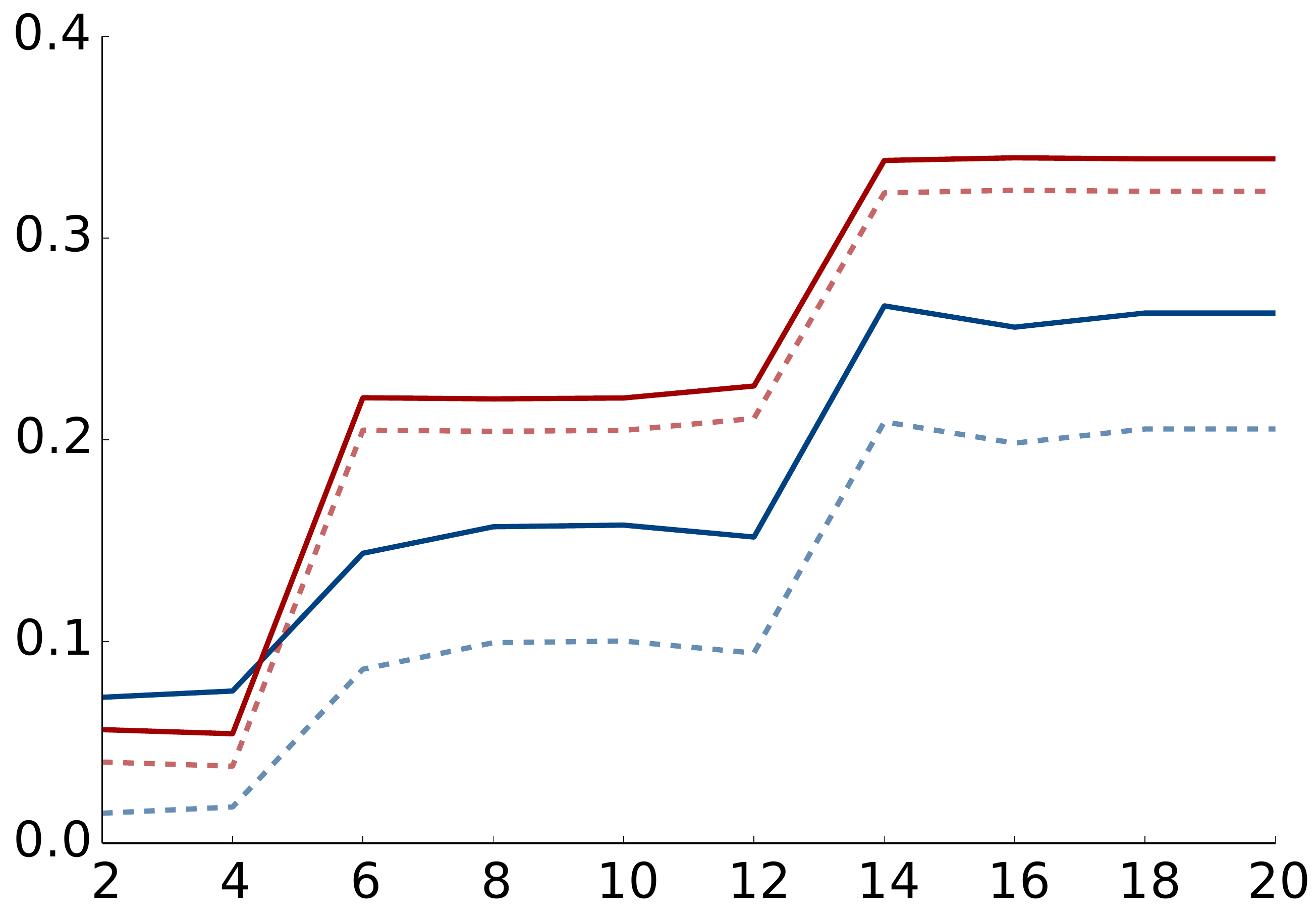}
		\caption{Keijzer-9}\label{fig:rie-eie-kei9}
	\end{subfigure}
	\hfill
	\begin{subfigure}[t]{0.3\textwidth}
		\centering
		\includegraphics[scale=0.2]{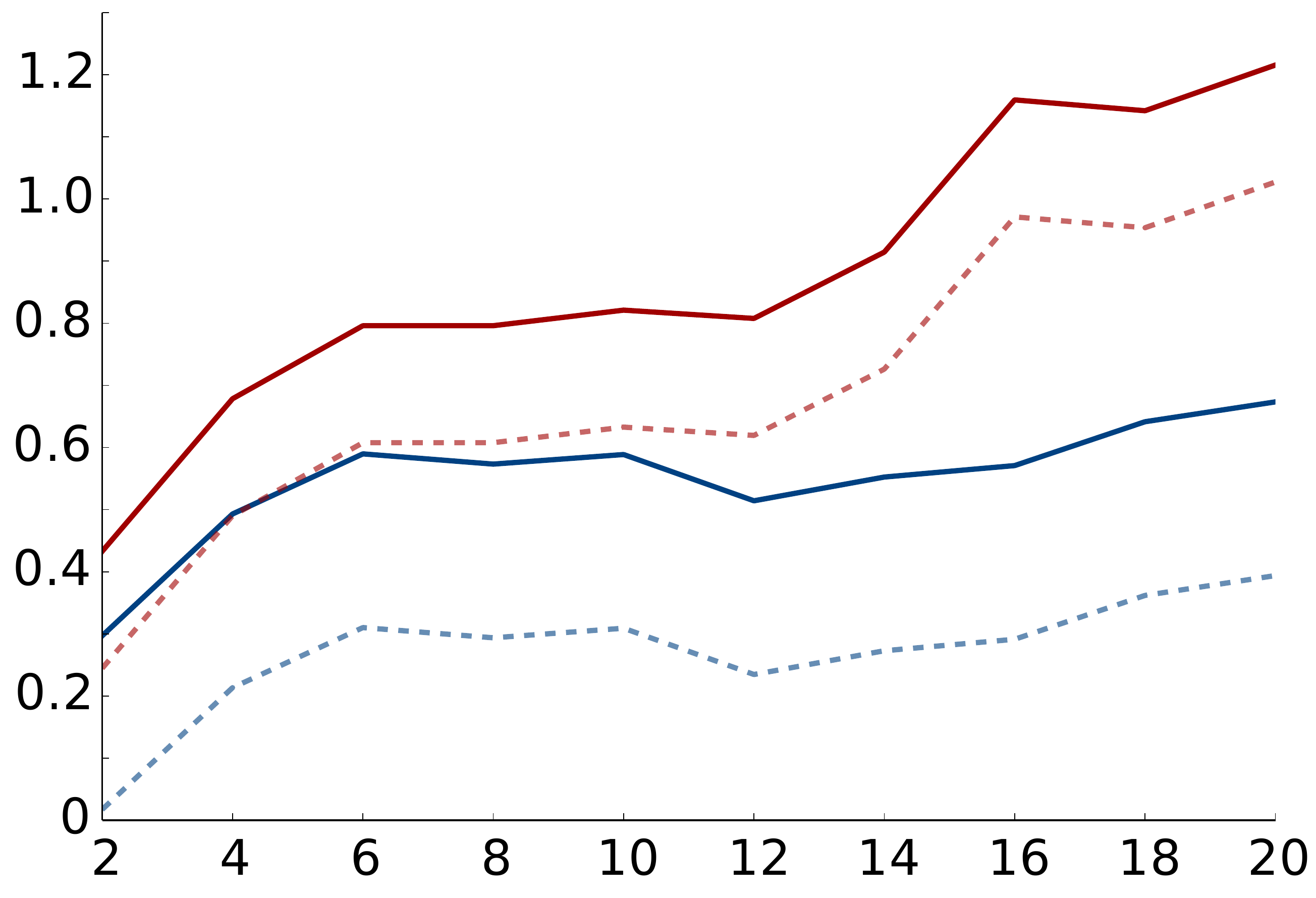}
		\caption{Vladislavleva-1}\label{fig:rie-eie-vla1}
	\end{subfigure}
	\\[0mm]
	\begin{subfigure}[t]{0.32\textwidth}
		\centering
		\includegraphics[scale=0.2]{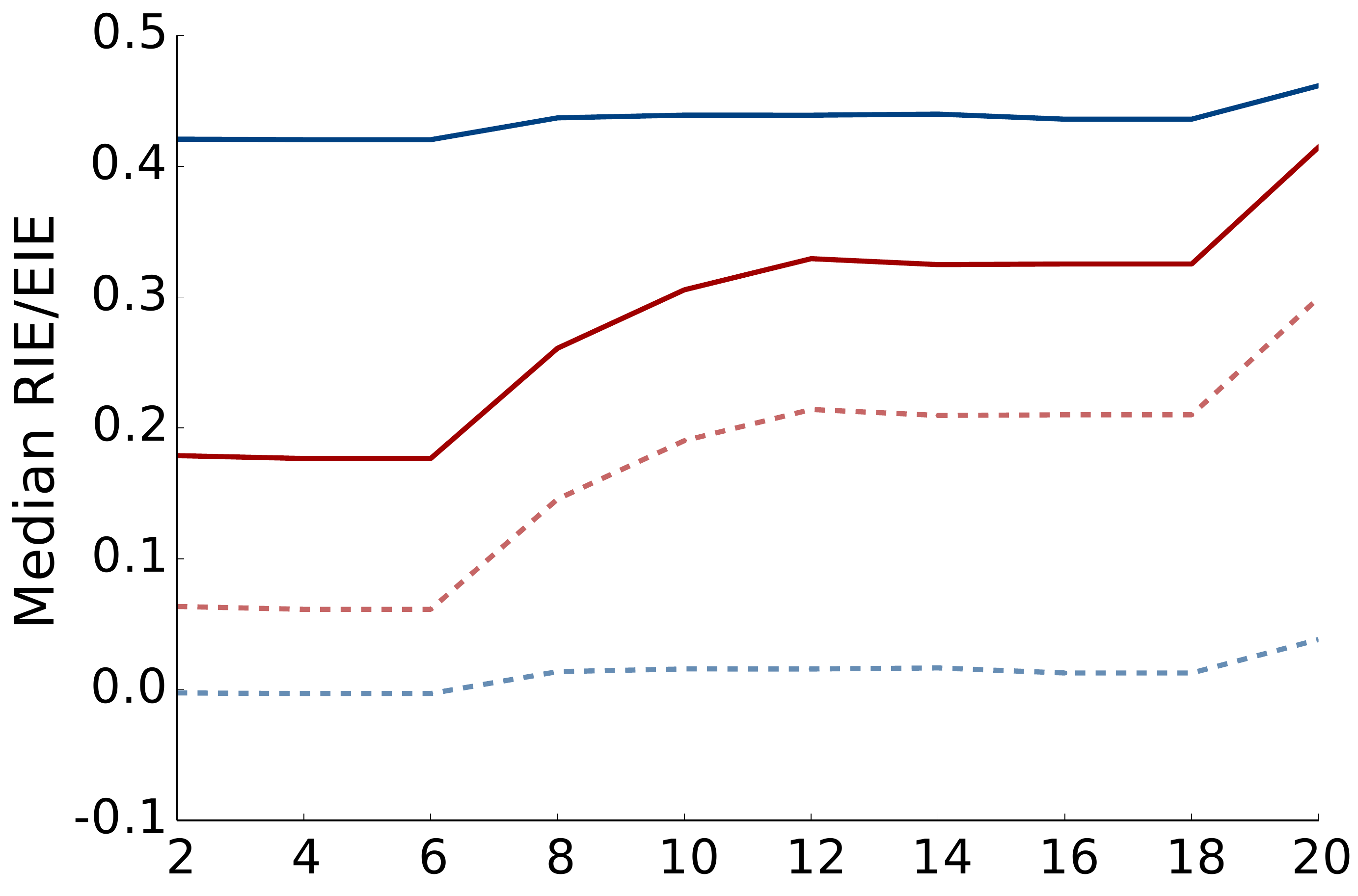}
		\caption{Vladislavleva-2}\label{fig:rie-eie-vla2}
	\end{subfigure}
	\hfill
	\begin{subfigure}[t]{0.3\textwidth}
		\centering
		\includegraphics[scale=0.2]{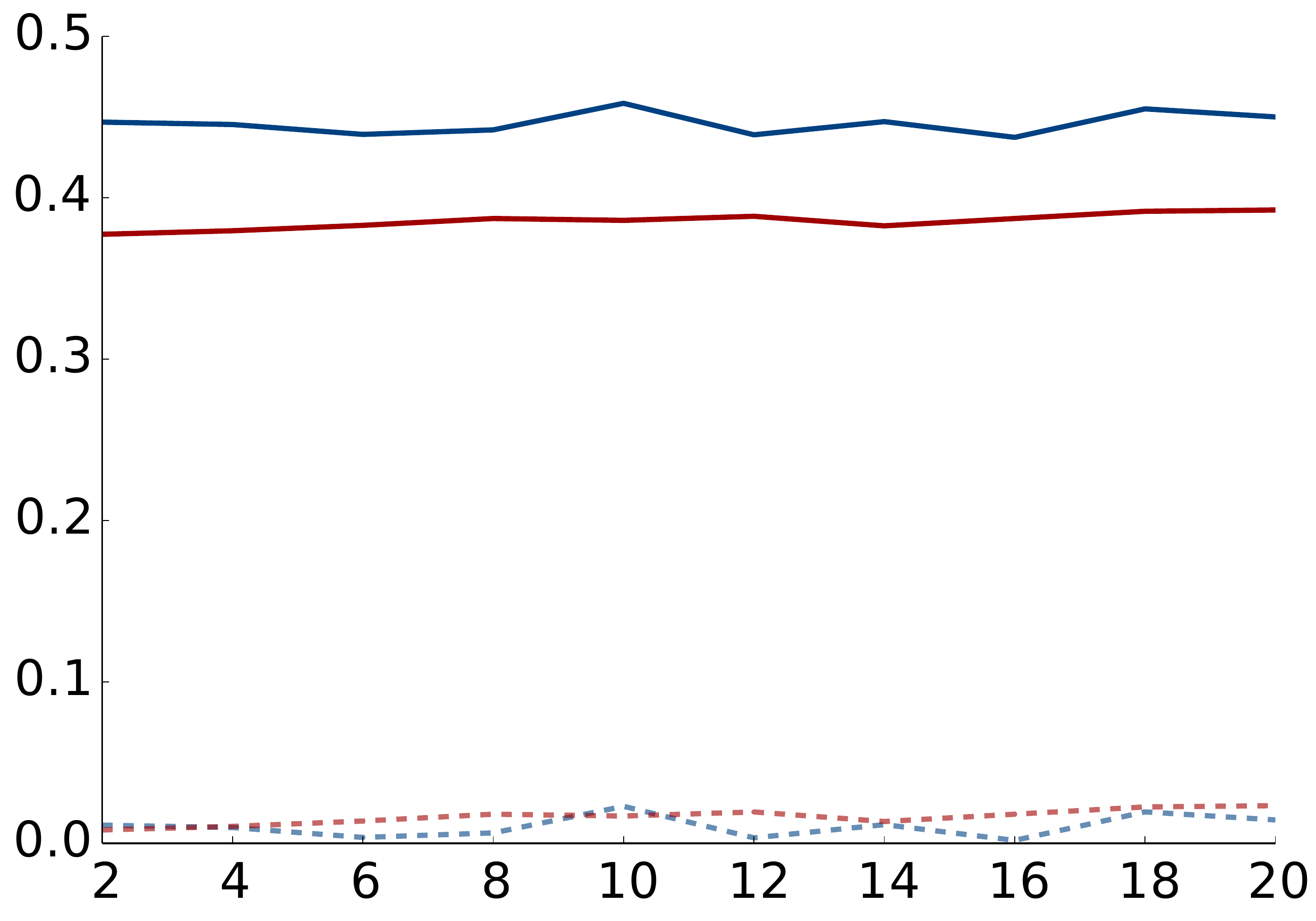}
		\caption{Vladislavleva-3}\label{fig:rie-eie-vla3}
	\end{subfigure}
	\hfill
	\begin{subfigure}[t]{0.3\textwidth}
		\centering
		\includegraphics[scale=0.2]{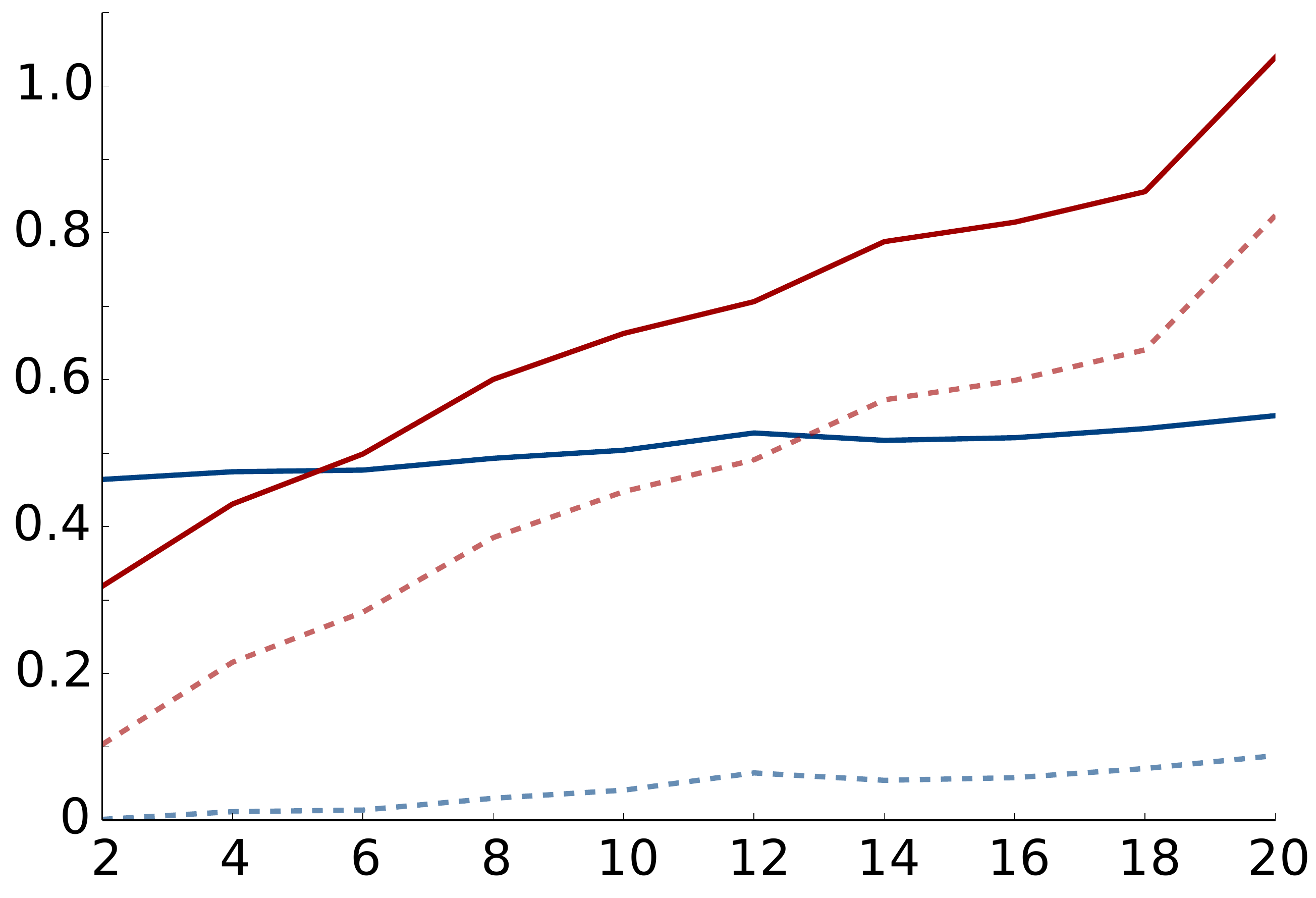}
		\caption{Vladislavleva-4}\label{fig:rie-eie-vla4}
	\end{subfigure}
	\\[0mm]
	\begin{subfigure}[t]{0.32\textwidth}
		\centering
		\includegraphics[scale=0.2]{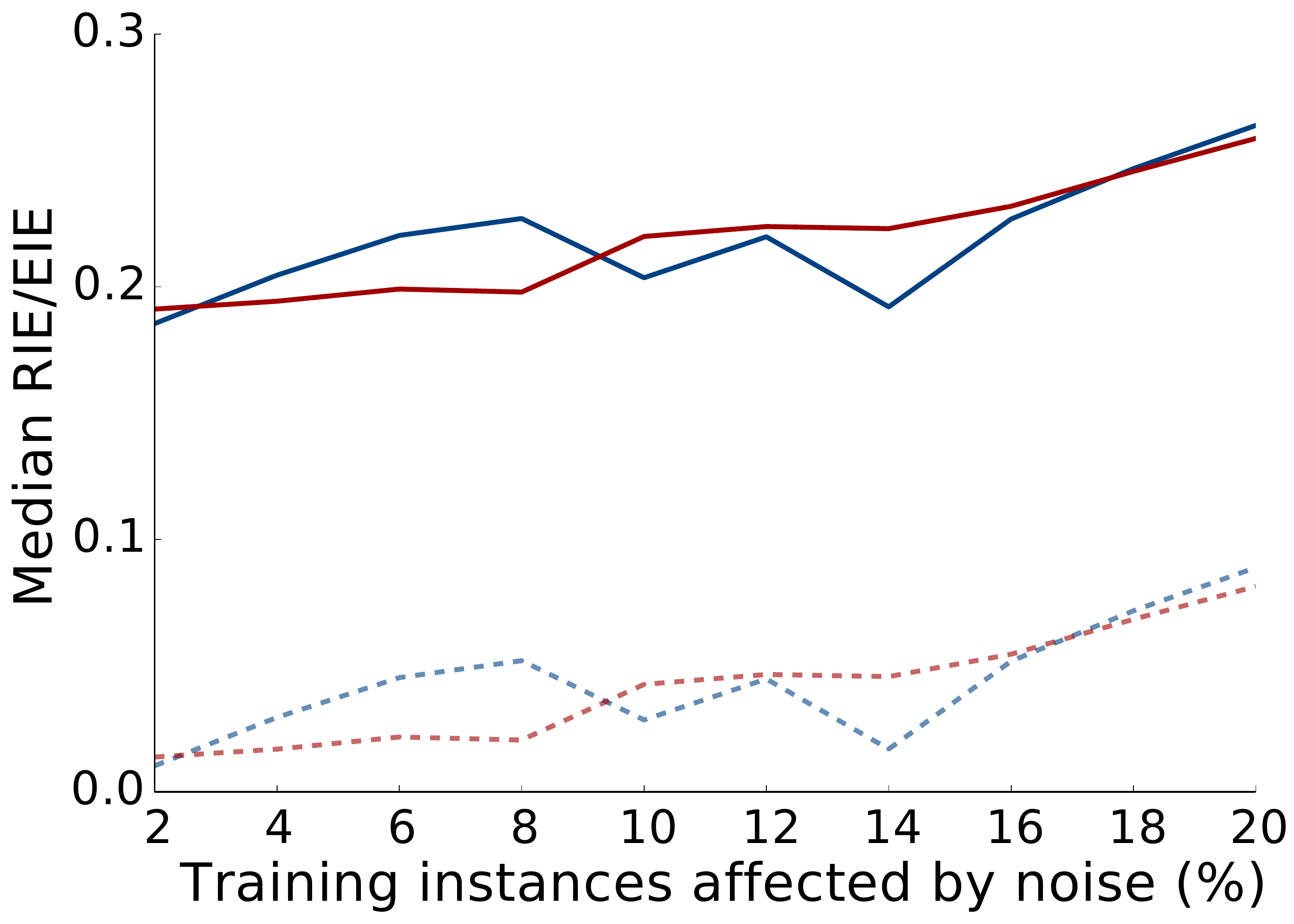}
		\caption{Vladislavleva-5}\label{fig:rie-eie-vla5}
	\end{subfigure}
	\hfill
	\begin{subfigure}[t]{0.3\textwidth}
		\centering
		\includegraphics[scale=0.2]{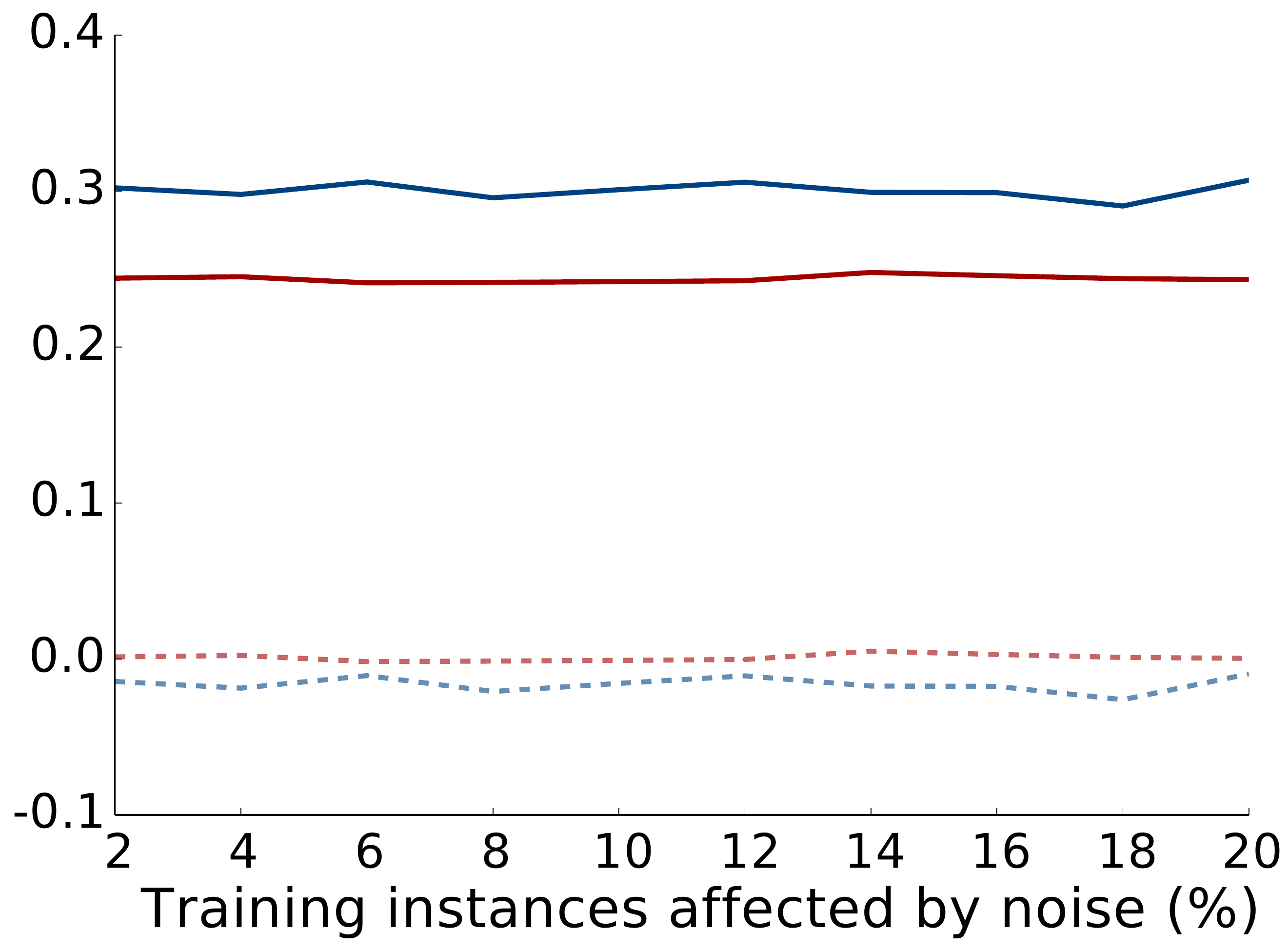}
		\caption{Vladislavleva-7}\label{fig:rie-eie-vla7}
	\end{subfigure}
	\hfill
	\begin{subfigure}[t]{0.3\textwidth}
		\centering
		\includegraphics[scale=0.2]{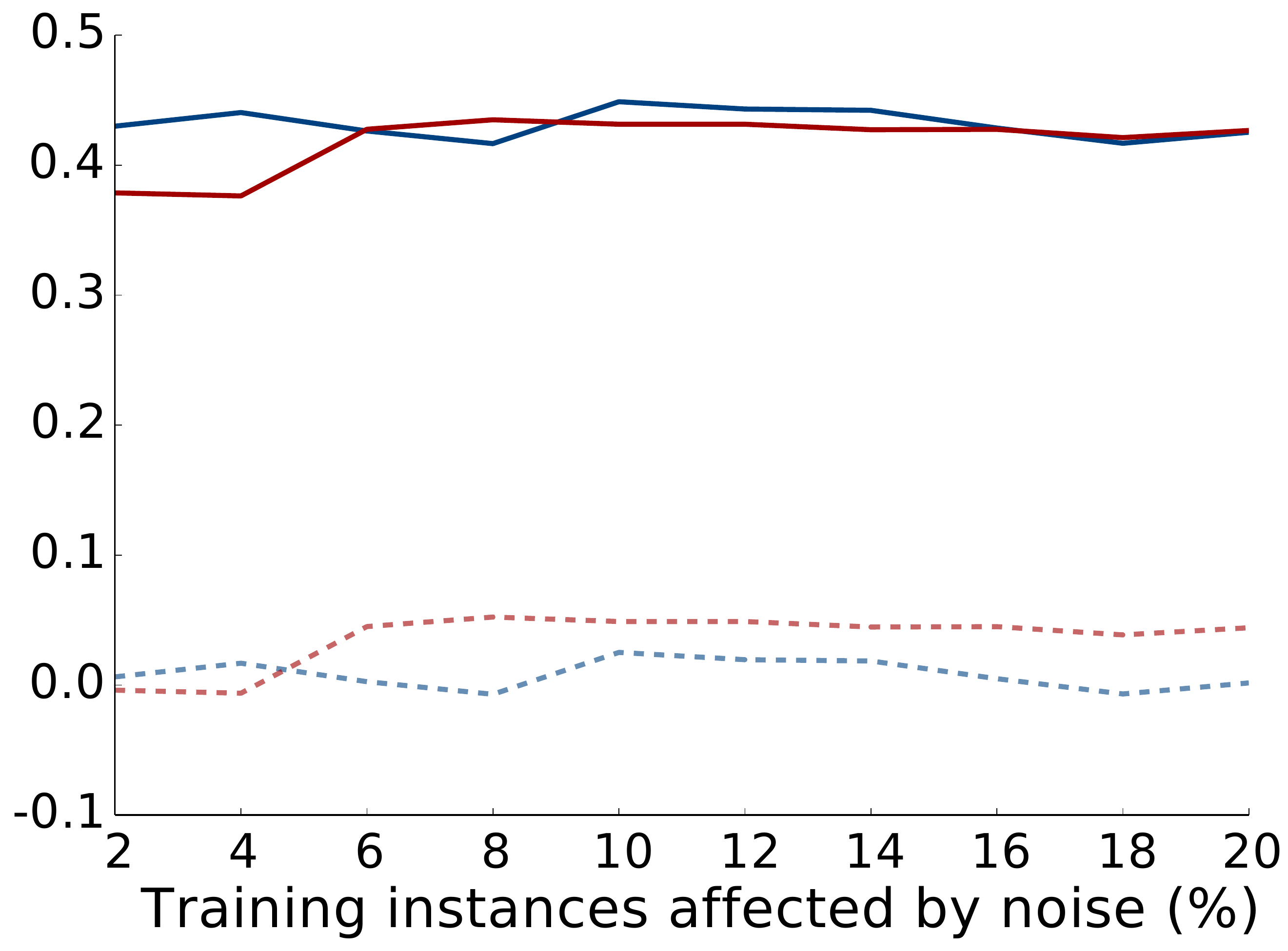}
		\caption{Vladislavleva-8}\label{fig:rie-eie-vla8}
	\end{subfigure}
	\caption{Median test RIE and EIE obtained by GP and GSGP for each dataset.}
	\label{fig:rie-eie}
\end{figure*}

Figure \ref{fig:rie-eie}, in turn, shows the median values for the EIE and RIE measures presented in Section \ref{sec:robustness-mestrics-regres}, obtained by GSGP and GP methods for different noise levels considering only the test set. When analyzing RIE values, we verify that GSGP is less robust to noise than GP for all noise levels in 10 datasets---Keijzer-2, Keijzer-3, Keijzer-4, Keijzer-6, Keijzer-7, Keijzer-9, Vladislavleva-1, Vladislavleva-2, Vladislavleva-4 and Vladislavleva-7---and for noise levels greater than or equal to 4\% for Keijzer-1 and Keijzer-8 and 6\% in the dataset Vladislavleva-8.

However, this scenario changes when we look at the values of EIE. GSGP is more robust than GP in all noise levels in six datasets---Keijzer-4, Keijzer-7, Keijzer-8, Vladislavleva-2, Vladislavleva-3 and Vladislavleva-7---and the opposite happens in only two datasets---Keijzer-6 and Vladislavleva-1. Besides, we can observe that GSGP obtains smaller EIE values than GP for noise levels smaller than $18\%$ in the datasets Keijzer-1 and Keijzer-3. On the other hand, GP outperforms GSGP in terms of EIE for noise levels greater than $4\%$ in the datasets Keijzer-9 and Vladislavleva-4. These analyses indicate that, overall, GSGP is more robust to noise than GP according to the EIE measure.

The main reason for these contradicting results lies on what these measures regard as important to quantify noise robustness. As presented in Section \ref{sec:robustness-mestrics-class}, the method performance in the dataset with no noise has very low influence in the RLA measure---and consequently in its regression counterpart (RIE). The ELA and EIE, on the other hand, add a term to their respective equations to represent the behavior of the model in the data without controlled noise. 
As GSGP performs better than GP in the majority of scenarios when no noise is present, it is natural that EIE considers it more robust to noise than RIE.


In order to compare the results presented in Figures \ref{fig:nrmse} and \ref{fig:rie-eie}, 
we conducted three paired one-tailed Wilcoxon tests comparing GP and GSGP under the null hypothesis that their median performance---measured by their median test NRMSE, RIE and EIE in all datasets---are equal. The adopted alternative hypotheses
differ according to to the overall results presented in Figures \ref{fig:nrmse} and \ref{fig:rie-eie}: GSGP outperforms GP in terms of NRMSE and EIE and GP outperforms GSGP in terms of RIE. The \textit{p}-values reported by the tests are presented in Table \ref{tab:results}. 
Considering a confidence level of 95\%, the symbol \eq indicates the null hypothesis was not discarded and the symbol \up (\down) indicates that GSGP is statistically better (worse) than GP. For the NMRSE measure, GSGP outperforms GP in datasets with $0\%$, $2\%$, $4\%$, $6\%$, $8\%$ and $12\%$ of noise. However, there are no statistical differences when the noise level is greater than $12\%$, which indicates that GSGP performance approximates from GP. When analyzing the robustness measures, RIE indicates that GP is more robust than GSGP in all noise levels. However, the same is not true for the EIE measure, which indicates GSGP is more robust than GP with low levels of noise ($2\%$ and $4 \%$) and have no significant differences for noise levels greater than $4\%$.


%% file: 6-conclusions.tex
\section{Conclusions}

This paper presented an analytic study of the impact of noisy data on the performance of GSGP when compared to GP in symbolic regression problems. The performance of both methods was measured by the normalized RMSE and two robustness measures adapted from the classification literature to the regression domain, namely Relative Increase in Error (RIE) and Equalized Increase in Error (EIE), in a test bed composed of 15 synthetic datasets, each of them with 11 different levels of noise equally spaced in $[0.00,0.20]$.

Results indicated that GP is more robust to all levels of noise than GSGP when the RIE measure is employed to analyze the outcomes. However, when the NRMSE or EIE values were analyzed, GSGP outperformed GP in terms of robustness to lower levels of noise and presented no significant differences regarding GP in higher levels of noise. Overall, these outcomes indicate that, although GSGP performs better than GP in low levels of noise, the methods tend to perform equivalently for larger levels of noise.
Given these conclusions, potential future developments include investigating techniques to identify the noisy instances in order to remove them or minimize their importance during the search.